\documentclass[pdflatex,sn-mathphys-num]{sn-jnl}% Math and Physical Sciences Numbered Reference Style
\usepackage{graphicx}%
\usepackage{multirow}%
\usepackage{amsmath,amssymb,amsfonts}%
\usepackage{amsthm}%
\usepackage{mathrsfs}%
\usepackage[title]{appendix}%
\usepackage{xcolor}%
\usepackage{textcomp}%
\usepackage{manyfoot}%
\usepackage{booktabs}%
\usepackage{algorithm}%
\usepackage{algorithmicx}%
\usepackage{algpseudocode}%
\usepackage{listings}%
\usepackage{subcaption}%

\newcommand{\RL}[1]{\textcolor{black}{#1}}
\newcommand{\REBUTTAL}[1]{\textcolor{black}{#1}}
\theoremstyle{thmstyleone}%
\theoremstyle{thmstyletwo}%

\theoremstyle{thmstylethree}%

\begin{document}

\title[Article Title]{Automatic Discovery of Disease Subgroups by Contrasting with Healthy Controls}


\author*[1,2]{\fnm{Robin} \sur{Louiset}}\email{robin.louiset@gmail.com}

\author[1]{\fnm{Edouard} \sur{Duchesnay}}\email{edouard.duchesnay@cea.fr}

\author[1]{\fnm{Benoit} \sur{Dufumier}}\email{benoit.dufumier@cea.fr}

\author[1]{\fnm{Antoine} \sur{Grigis}}\email{antoine.grigis@cea.fr}

\author[1,2]{\fnm{Pietro} \sur{Gori}}\email{pietro.gori@telecom-paris.fr}

\affil[1]{\orgdiv{NeuroSpin}, \orgname{Université Paris-Saclay}, \orgaddress{\street{CEA}, \city{Gif-sur-Yvette}, \postcode{91191}, \country{France}}}

\affil[2]{\orgdiv{LTCI}, \orgname{Institut Polytechnique de Paris}, \orgaddress{\street{Télécom Paris}, \city{Palaiseau}, \postcode{91120}, \country{France}}}


\abstract{In biomedical Subgroup Discovery, practitioners are interested in discovering interpretable and homogeneous subgroups within a group of patients. In this paper, assuming that healthy subjects (i.e., controls)  share common but irrelevant factors of variation with the patients, we motivate and develop a Contrastive Subgroup Discovery method, entitled Deep UCSL. By contrasting patients with controls, Deep UCSL identifies subgroups driven solely by pathological factors, ignoring common variability shared with healthy subjects. Our framework employs a deep feature extractor to learn a discriminative representation space. Mathematically, we derive a novel loss based on the conditional joint likelihood of latent clusters and patient/control labels, optimized via an Expectation-Maximization strategy alternating between subgroup inference and feature encoder updates. A regularization term further encourages representations to capture disease-specific variability while ignoring variability shared with controls. Compared to previous related works, our approach quantitatively improves the quality of the estimated subgroups, as demonstrated on a MNIST example and four distinct real medical imaging datasets. Code and datasets are available at: \url{https://github.com/rlouiset/deep_ucsl}.}




\maketitle

\section{Introduction}\label{sec1}
\label{sec:introduction}

%% 1 - Clustering is interesting but is based on the general variability
\noindent In the past decades, unsupervised and self-supervised learning techniques have proven to be particularly effective at identifying relevant patterns and factors of variation within a dataset. Combined with powerful Neural Networks (NNs), these methods can produce semantically rich representations \cite{chen_simclr_2020},\cite{he_2020}, \cite{zheng_2021}. Notably, unsupervised Deep Clustering (DC) methods \cite{caron_2018},\cite{caron_2020},\cite{li_2021},\cite{van_gansbeke_2021} seek to produce a suitable representation space for identifying homogeneous latent clusters based on the \textit{general variability} of the entire dataset (i.e., imaging patterns common to all samples).
%, i.e. the features shared across all images. % aim at learning a suitable representation space for the identification of homogeneous latent clusters based on the \textit{general variability} of the entire data-set (i.e., imaging features common to all samples). 

%% 2 - Subgroup Discovery is better because it is based on the specific variability, intuitive examples of dermatology and MNIST
\noindent With a different perspective, Subgroup Discovery (SD) in medical applications \cite{atzmueller_2015}, \cite{klosgen_1996}, \cite{yang_subtypes_2021} aims at identifying relevant latent subtypes/subgroups that arise from the \textit{pathological variability} of the diseased population and not from the irrelevant common variability that may exist in both healthy subjects (i.e., controls) and diseased patients. For instance, in dermatology, practitioners are interested in identifying a stratification within a dataset of malignant melanomas. In this case, the objective would be to retrieve relevant dermatological subgroups, from a medical standpoint (\textit{e.g.,} nodular/lentigo melanoma), for a more targeted drug or treatment delivery. In this paper, we argue that the discovery of these subgroups should be based on disease-specific patterns (e.g.: texture, color, asymmetry) rather than irrelevant patterns shared with healthy skin samples (e.g.: skin color, hair, moles or nevi). 

\begin{figure}[ht!]
    \centering
    \includegraphics[width=0.9\textwidth]{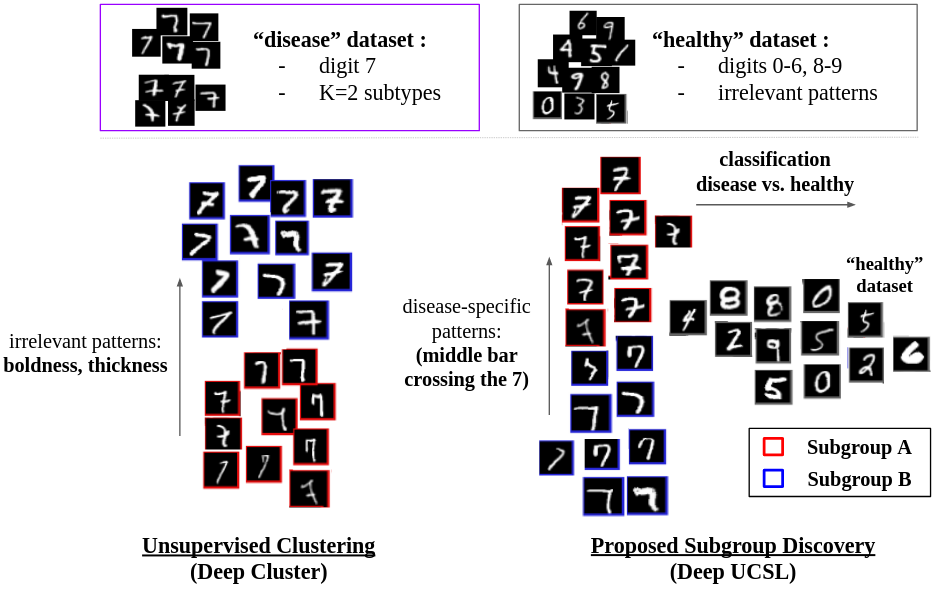}
    \caption{Comparison between a Deep Clustering method (Deep Cluster \cite{caron_2018}) and the proposed Subgroup Discovery method (Deep UCSL) on a subtype discovery task within the digit 7. The relevant subtypes stem from the digit-7's specific variability, \textit{i.e.,} 7 with a middle-cross bar (red) and without (blue). We show two 2D PCA plots of the representation spaces learnt by the two methods. Deep Cluster is driven by the general variability of the digits, in particular the boldness. Differently, Deep UCSL uses a supervised classification task to contrast with the ''healthy'' class (\textit{i.e.,} the rest of the digits). In this way, it encourages the ”disease” subtype identification to discard the shared factors of variability, focusing only on the relevant variability of the 7 (\textit{i.e.,} the middle crossbar).
    }
    \label{fig:deep_ucsl_compared_with_deep_cluster}
\end{figure}

\noindent  In Fig.~\ref{fig:deep_ucsl_compared_with_deep_cluster}, we use an intuitive toy example based on the MNIST dataset to better clarify the differences between Deep Clustering and Subtype Discovery. We consider the digit ''7'' as the pathological group and all the other digits as the healthy group. Results show how Deep Cluster's subgroups \cite{caron_2018} of the digit ''7'' are only defined by the most predominant characteristics (i.e.: boldness of the digit) common to all digits.
%, which are shared with the rest of the digits. 
Instead, the proposed Subtype Discovery method, called Deep UCSL, disregards these common characteristics and uses only the specific patterns of the digit ''7'' (i.e., the presence of the crossing middle bar) to define the subgroups.

%% 3 - Subgroup Discovery with ML is important to push forward the research in many medical domains
\noindent In many medical domains, the discovery of new subgroups is left to human experts. However, this task may be difficult and observer-dependent, with potentially high-dimensional images and subtle imaging patterns. Furthermore, there are often no precise guidelines to define %which imaging patterns are relevant
the subgroups, and there is also considerable disagreement among pathologists, with high intra- and inter-rater variability \cite{yang_subtypes_2021}. 
%These reasons 
This motivates the need 
%to develop 
for new machine-learning methods to help human experts discover or validate subgroups, while relying on reproducible, data-driven, and objective imaging patterns.

\section{Related works}

\subsection{Subgroup Discovery applications}
%% 5 - Related Fields of applications
In the last decade, Subgroup Discovery works have been proposed across a variety of topics. In medicine, a lot of works have been conducted to refine the characterization of cancer variants \cite{carey_2006}, \cite{planey_2016}. 
In psychiatry, mental diseases are known as being extremely heterogeneous, and multiple works have tried to refine the disease categorization into homogeneous subgroups \cite{chand_2020}, \cite{ferreira_2017}, \cite{honnorat_2019}, \cite{yang_probing_2021}, \cite{wen_2020}, \cite{louiset_2023}, \cite{louiset_separating_2024}. However, most of these subgroup discovery methods either rely on priors dependent on the domain (\textit{e.g.,} the aspect of pathological patterns of neurodegenerative disorders), or are based on user-defined features and linear predictors. Notably, emerging machine learning methods, such as  HYDRA~\cite{varol_2017} and our preliminary work UCSL~\cite{louiset_2021}, have proposed a framework for Subgroup Discovery, where the supervised classification (healthy vs. diseased samples) and the subgroup identification (clustering of diseased samples only) depends on each other. This framework respects two properties: 1) subgroups should be identified from \textit{diseased} samples \textit{only}, and 2) pathological subgroups should be correctly discriminated from the healthy class in the estimated representation space. However, these works are limited because they require a user-defined feature extraction step and are constrained to linear predictors only. Indeed, each inferred pathological subgroup is being discriminated from the healthy class with a linear classifier. To sum up, actual subgroup discovery works are either not generalizable to other domains or depend on user-defined feature choices and generally do not leverage the representation quality of non-linear deep learning methods. \\

\subsection{Unsupervised Deep Clustering}
%% 7 - Deep Clustering does not focus on specific variability
%
Recent works \cite{caron_2018}, \cite{caron_2020}, \cite{lv_2021}, \cite{zhan_2020}, proposed to uncover semantically relevant groups of samples in a dataset without using other auxiliary tasks. Notably, Deep Cluster \cite{caron_2018} alternates between the pseudo-label estimation phase (i.e.: the clustering estimation step) and the update of encoder parameters. At each epoch, previous clustering assignments produced by a K-Means method are used as pseudo-labels for minimizing a Cross-Entropy loss with the network output logits. A major challenge is clustering degeneracy, where imbalanced assignments cause most samples to collapse into a few clusters. To mitigate this, Deep Cluster \cite{caron_2018} and ODC \cite{zhan_2020} proposed to weigh samples' importance with the inverse of the associated cluster size. Another work, SwAV \cite{caron_2020}, regularizes the prototype centroids so that they respect the equipartition constraint (i.e.: samples get equally distributed across the clusters). 
Another issue is cluster identity inconsistency across epochs: clusters reshuffle when K-Means is recomputed, making the classifier weights outdated. To address this, \cite{caron_2018} reinitializes the classifier head each epoch, \cite{zhan_2020} jointly updates centroids and classifier weights, and \cite{caron_2020} replaces the classifier with learnable prototypes.

\subsection{Self-Supervised Learning.}
%\label{contrastive}
%% 7 - Contrastive Learning may reduce general variability up to a certain point
Another relevant line of work comprises self-supervised, and particularly contrastive, learning methods \cite{chen_simclr_2020}, \cite{grill_2021}, \cite{he_2020}, \cite{zheng_2021}, \cite{barbano_unbiased_2023}, \cite{dufumier_integrating_2023} which learn representations potentially suitable for downstream tasks, such as clustering or classification. A pivotal method of this literature is SimCLR \cite{chen_simclr_2020}, which encourages the encoder to be invariant to user-defined image augmentation (\textit{i.e.,} it discards data-augmentation variability), while repelling views of different images. 
Contrastive learning methods only align representations of the same image, which implies that several images with the same semantic content can be pulled apart. This flaw is known as the Class Collision problem \cite{zheng_2021}, \cite{dufumier_conditional_2021} and may be harmful to downstream clustering and classification tasks. To solve this problem, two lines of works have emerged. In Supervised Contrastive Learning (SupCon) \cite{khosla_2020}, authors use a loss where positive and negative samples are defined based on their class (e.g. healthy or patient).\\ 
The second methods use a contrastive loss within a Deep Clustering or Deep Nearest Neighbor \cite{li_2021}, \cite{caron_2020}, \cite{tsai_2021}, \cite{van_gansbeke_2021}. For example, PCL \cite{li_2021} proposes using cluster assignments to align positive views from the same cluster rather than from the same image. Other works, such as \cite{dang_2021, van_gansbeke_2021}, propose aligning nearest neighbors instead. These works either leverage the disease/healthy label information (first group) or encourage clustering structures (second group). However, a Subgroup Discovery method should combine both properties to effectively define homogeneous groups \textit{only} among diseased samples and \textit{only} based on the pathological patterns/variability of the disease.  

\noindent \RL{Recently, self-supervised learning works based on masked image modeling \cite{he_2021}, \cite{fang_2023}, \cite{xie2022simmimsimpleframeworkmasked}, with Transformers encoder \cite{dosovitskiy_2021} demonstrated that reconstructing randomly masked patches yields powerful semantic representations, even in medical imaging such as retinal imaging \cite{zhou_2023}, and chest X-Ray imaging \cite{xiao_2023}, \cite{yao_2024}. However, these methods do not cluster samples according to latent disease-related subgroup variability and they do not encourage the representations to discard shared patterns between healthy and diseased subjects (they generally encourage capturing it instead, using masked reconstruction). Thus, these approaches do not explicitly model disease-specific subgroups. \\
\noindent Vision–language models \cite{radford_learning_2021} improved transferability by leveraging paired image–text data. However, their supervision in medical imaging \cite{wang2022medclipcontrastivelearningunpaired} reflects existing clinical ontologies and disease labels, making them suited to recognizing known concepts rather than discovering novel pathological subgroups. They are therefore not intended to uncover latent disease stratifications beyond what is already encoded in text.}

%%%%%%%%%%%%%%%%%%%%%%%%%%%%%%%%%%%%%%%%%%%%%%%%%%%%%%%%%%%%%%%%%%%%%%%%%%%%%%%%%%%

\section{Contributions}
\label{sec:contributions}
%% 4 - Subgroup Discovery methods exist, but they are limited to linear patterns so we propose a deep learning alternative
\noindent To overcome these shortcomings, this paper proposes a Deep Unsupervised Clustering driven by Supervised Learning (Deep UCSL), where we use a Deep Neural Network as an automatic feature extractor to produce representations suitable for identifying relevant disease subgroups that are contrasted from the healthy class. As in UCSL \cite{louiset_2021}, we derive our objective by seeking the parameters that maximize the conditional joint likelihood between latent subgroups and supervised labels. However, as a difference, the feature space is estimated with a trainable deep encoder that allows an automatic non-linear feature extraction. Then, as in UCSL \cite{louiset_2021}, we propose the use of an Expectation-Maximization optimization process to alternate between the conditional joint likelihood parameters update and the subgroups pseudo-labels estimation. However, here we propose a new regularization loss that improves the separation between healthy and diseased patterns and, differently from UCSL, it 
%make our optimization procedure to 
guarantees the monotonical convergence of the model parameters.\\
In our approach, the optimization procedure consists of alternatively 1) estimating the subgroups pseudo-labels (i.e.: pathological subgroups within the disease classes) and 2) updating the features encoder. Our objective is to: a) correctly discover the pathological subgroups, b) encourage healthy samples not to belong to a pathological subgroup, and c) accurately discriminate each subgroup from the healthy class (Mixture-of-Classifying Experts).\\
To demonstrate the superiority of our method, we compare it with other state-of-the-art representation learning methods (i.e., Deep Clustering and Contrastive learning methods) on three different tasks: 1) 7-digit subgroup identification, 2) psychiatric application, 3) pneumonia subgroups identification, and 4) eye pathological subgroups identification. Deep UCSL substantially outperforms all other methods in these tasks. 
In a nutshell, our contribution is three-fold:
\begin{enumerate}
    \item To the best of our knowledge, we propose the first Deep Learning method for Subgroup Discovery that performs disease/healthy classification while identifying subgroups in the diseased class.
    \item We motivate and design a clustering regularization loss that forces the learned representation to disregard the healthy population variability focusing only on the disease-specific variability.
    \item A fair and careful evaluation of our method, as well as a comparison with recent state-of-the-art methods.
\end{enumerate}

\section{Methodology}
In this section, we will introduce the theoretical background of Deep UCSL and explains our assumptions and choices concerning its implementation in practice. From a general comprehensive point of view, Deep UCSL operates in two phases: (1) training on a dataset with known labels (0: healthy, 1: disease), typically determined by medical experts, and (2) inference, where the model predicts whether a new sample is healthy or diseased and predicts probabilistic assignment over a set of $K$ learned subgroups (pathological subgroups'centroids are estimated and discovered during the training process, without subgroup-level supervisions). As commonly done in clustering-like methods, the number of subgroups $K$ is chosen by the user. \REBUTTAL{In practice, $K$ can be guided by domain expertise or classical model selection criteria such as the silhouette score~\cite{rousseeuw1987silhouettes}, the Davies--Bouldin index~\cite{davies1979cluster}, or information-theoretic criteria (e.g., BIC). Extensions such as X-means~\cite{pelleg2000xmeans} provide principled approaches to estimate $K$ by optimizing model complexity. In the deep clustering literature, \citet{xie2016dec} further propose using the ratio of training to validation loss as an overfitting indicator: a sharp decrease signals that $K$ is too large. In our setting, the mixture-of-experts formulation provides a natural verification, as inappropriate values of $K$ typically degrade classification performance, which can serve as a practical proxy for model selection.}

\subsection{Mathematical formulation}

\begin{figure*}[!tbp]
    \centering
    \includegraphics[width=0.99\textwidth]{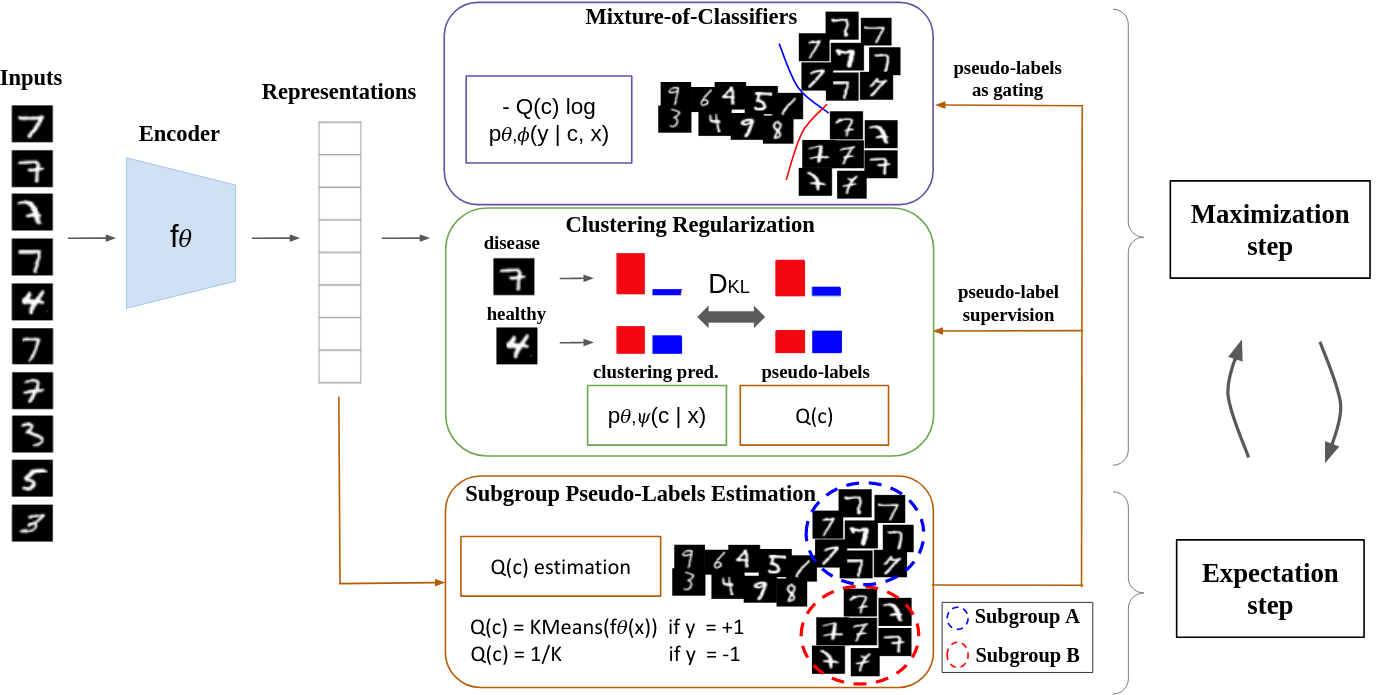}
    \caption{A schematic diagram of Deep UCSL with $K=2$ subgroups (red and blue). At each epoch, K-Means  produces subgroup pseudo-labels during the Expectation step (in brown). These pseudo-labels are then used to weight a classification Mixture-of-Experts (in purple) between the ''healthy'' class (digits 0-6, 8-9) and the ''disease'' class (digit 7). Additionally, the pseudo-labels are also used for the clustering regularization (in green),
    %, as in Deep Cluster \cite{caron_2018}. 
    where  uniform pseudo-labels (i.e.: $\frac{1}{K}$) are used to regularize the healthy class distribution, so that healthy samples are equidistant from all the diseased subgroups. This forces the learnt representation  to disregard the general variability, common to both healthy and diseased samples.}
    \label{fig:deep_ucsl}
\end{figure*}

\noindent Let $(X,Y)=\{(x_i,y_i) \}_{i=1}^N$ be a labeled dataset composed of $N$ samples. We will restrict to the binary (e.g., patient/control) classification paradigm, $y_i \in \{-1,+1 \}$, which is very common in medical imaging. We will denote with $N^+$ and $N^-$ ($N=N^+ + N^-$) the number of positive and negative samples, respectively. Our objective is to estimate the latent pseudo-labels\footnote{we call them pseudo-labels, since we assume that subgroups labels are not known at training} of subgroups within disease samples ($y_i=+1$). The membership of each sample $i$ to latent subgroups is modeled via a latent categorical variable $c_i \in C=\{1,...,K\}$, where $K$ is the number of subgroups. We look for a discriminating model that maximizes the joint conditional likelihood:
\begin{equation}
    \sum_{i=1}^n \log p(y_i | x_i) = \sum_{i=1}^n \log \sum_{k=1}^K p(y_i, c_i=k | x_i)
    \label{eq:1}
\end{equation}
\noindent
To attain the three objectives (\textit{a), b), c)}) described in Sec.~\ref{sec:contributions}, we need to optimize  Eq. \ref{eq:1} with respect to both $p(c_i | x_i, y_i)$ and  $p(y_i | x_i, c_i)$. Indeed, we need to identify the subgroups only within the diseased samples (thus knowing $y$) and to accurately discriminate the healthy class from each subgroups (thus knowing $c$).
However, developing the joint conditional likelihood in Eq.~\ref{eq:1} would result in either $p(c_i | x_i,y_i)$ or  $p(y_i | x_i,c_i)$, but not in both. To solve that, as in UCSL \cite{louiset_2021}, we introduce a probability distribution $Q$ over the subgroups $C$, so that $\sum_{k=1}^K Q(c_i=k)=1 \quad \forall i$. To do so, we multiply the term inside the log by $Q(c_i=k)$ at the numerator and the denominator. And then, we use the Jensen inequality to pull $Q(c_i=k)$ at the numerator out of the log, to obtain a tractable, lower bound of Eq.~\ref{eq:1}:
\begin{equation}
    \begin{split}
        \sum_{i=1}^n \log p(y_i | x_i) =  \sum_{i=1}^n \log \sum_{k=1}^K Q(c_i=k)\frac{p(y_i, c_i=k | x_i)}{Q(c_i=k)} \geq \sum_{i=1}^n \sum_{k=1}^K Q(c_i=k) \log \left( \frac{p(y_i, c_i=k | x_i)}{Q(c_i=k)} \right)
    \end{split}
    \label{eq:2}
\end{equation}
\noindent According to the Jensen's inequality, the equality holds when the distribution $Q(c_i=k)$ is equal to $p(c_i=k | x_i, y_i)$:
\begin{equation}
    Q(c_i=k) = \frac{p(y_i, c_i=k | x_i)}{ \sum_{k=1}^K p(y_i,c_i=k | x_i)} = \mathbf{p(c_i=k | x_i, y_i)}
    \label{eq:3}
\end{equation}
\noindent Then, Eq.~\ref{eq:2} can be rewritten with respect to both $p(y_i | x_i, c_i)$ and $Q(c_i)$ (estimated to approximate $p(c_i | x_i, y_i)$):
\begin{equation}
    \begin{split}
         \sum_{i=1}^n \log p(y_i | x_i) \geq & \underbrace{ \sum_{i=1}^n \sum_{k=1}^K Q(c_i=k) \log  \mathbf{p(y_i |x_i, c_i=k)} }_{\text{Mixture-of-Classifying Experts term}} - \hspace{-0.5cm} \underbrace{D_{KL}(Q(c)||p(c | x)) }_{\text{\parbox{4cm}{\centering Clustering Regularization \\[-4pt] (always non-negative)}}}
     \label{eq:4}
    \end{split}
\end{equation}
%We use the lower bound Eq.~\ref{eq:3} as proxy for maximizing Eq.~\ref{eq:1} which is intractable in practice. 

\noindent Our goal is to learn a single representation space where both the classifying experts $p(y_i | x_i, c_i=k)$ and the disease subgroup $p(c_i=k |x_i, y_i=+1)$ can be accurately estimated. To this end, we propose using a deep encoder $f_\theta$ with parameters $\theta$ for feature extraction and two neural networks with parameters $\phi$ and $\psi$ for the classifying experts $p_{\theta,\phi}(y_i | c_i=k, x_i)$ and the unsupervised clustering head $p_{\theta,\psi}(c_i=k | x_i)$, respectively. An overview of the proposed method can be seen in Fig.\ref{fig:deep_ucsl}. To sum up, we train a mixture of classifiers using the class/groups labels $y$ (supervised learning) to guide the discovery of subgroups without using subgroup labels $c$ (unsupervised clustering). From a mathematical point of view, we optimize the cost function (Eq.~\ref{eq:4}) using an Expectation-Maxization (EM) optimization where we first estimate $Q$ as $p(c_i=k |x_i, y_i)$ and then $p(y_i | x_i, c_i=k)$ and $p(c_i=k | x_i)$. No subgroups labels $c$ are needed, only the class labels $y$. The EM optimization process alternatively:
\begin{enumerate}
    \item estimates $Q$ as $p(c_i=k | x_i, y_i)$ (E-step, Eq.~\ref{eq:3}) at the end of each epoch, freezing the encoder $f_\theta$ 
    \item freeze $Q$, update the encoder $f_\theta$, by estimating $p(y_i | x_i, c_i=k)$ and $p(c_i=k | x_i)$ batch-wise and maximizing Eq.~\ref{eq:4} (M-step)
\end{enumerate}

\subsection{Comparison with UCSL}
%This mathematical framework is similar to the one of UCSL \cite{louiset_2021}, but with important differences. First, it makes use of a deep features encoder instead of linear models. Second, we do not assume that $p_\theta(c|x) = Q(c)$, as in UCSL, but we force it by explicitly introducing and minimizing the clustering regularization term $KL(Q(c) || p_{\theta,\psi}(c|x_i))$. This guarantees the monotonic convergence of the optimization procedure, which was not the case in UCSL. Third, since we want to estimate subgroups only within disease samples ($y_i=+1$), all healthy samples are assigned a uniform probability for all subgroups.
%Contrarily to UCSL, we guarantee the convergence of the optimization procedure without assuming that $p_\theta(c_i|x_i) = Q(c_i)$,  but forcing it by explicitly introducing and minimizing the clustering regularization term $KL(Q(c_i) || p_{\theta,\psi}(c_i|x_i))$. (See convergence Proof in the Appendix). This means that $p_\theta(c_i|x_i)$, the estimated clustering distribution, is not simply extended to all samples regardless their label $y$, as in UCSL, but the representation space is estimated so that the unsupervised clustering  $p_\theta(c_i|x_i)$ gives the same result as the \virg{supervised} clustering $p(c_i=k | y_i, x_i)$, namely knowing the label $y$. As explained in Sec.\ref{sec:maximization}, this entails an encoder $f$ and a representation space where the general variability, common to both healthy and disease samples, is removed and the subtype estimation only depends on the specific variability of the disease class.
\noindent This mathematical framework is close to the one of UCSL \cite{louiset_2021}, but still differs. \\ 
\textbf{First}, Deep UCSL uses a deep feature encoder, instead of user-defined features, and two neural-networks for classification and subgroup estimate, instead of linear models. The key advantage of incorporating deep neural networks is the ability to learn complex, high-dimensional feature representations directly from raw data, rather than relying on predefined features. This improves the expressiveness and discriminative power of the learned representations, leading to more robust subgroup discovery. \\
\textbf{Second}, differently from UCSL \cite{louiset_2021}, we do not assume that $p_\theta(c|x) = Q(c)$, but we force it by explicitly introducing and minimizing the clustering regularization term $KL(Q(c) || p_{\theta,\psi}(c|x_i))$. \REBUTTAL{Indeed, although a similar KL term appears in UCSL, UCSL effectively neglects this term in the optimization by setting ($p(c|x)\approx Q(c)$). In other words, UCSL makes a strong assumption on the clustering distribution in the M-step so that $D_{KL}(Q|p_\theta(c|x))$ is zero. In contrast, our method explicitly optimizes $D_{KL}(Q|p_\theta(c|x))$, ensuring monotonic convergence (see Appendix A). To sum up,} this term was neglected in UCSL \cite{louiset_2021} and its inclusion is important since it allows the monotonic convergence of the optimization procedure.\\
\textbf{Third}, since we want to estimate subgroups only within diseased samples, all healthy samples are assigned a uniform probability for all subgroups. This strategy, also not proposed in UCSL, encourages the features encoder $f_\theta$ to produce a representation space where healthy samples do not belong (in terms of probabilistic assignment) to subgroups. \REBUTTAL{While HYDRA [37] imposed uniform cluster priors, UCSL’s authors \cite{louiset_2021} found that it was prone to yield collinear hyperplanes in the linear case, thus degrading the resulting clustering. In our deep model, we show that enforcing uniform cluster assignment for healthy controls acts as a useful regularizer. It forces the encoder to ignore shared variability between healthy and diseased samples, thereby structuring the representation by disease-specific
factors.} To enforce this assumption in our representation space, we incorporate this principle into the \textbf{clustering regularization term} (second term of Eq.~(4)). This term encourages \textbf{all} healthy samples to be uniformly distributed across the $K$ pathological subgroups. The second term represents the Kullback-Leibler (KL) divergence between $Q(c_i)$ and $p_{\theta}(c_i \mid x_i)$. This implies:
\begin{itemize}
    \item The lower bound is tight when $Q(c_i) = p(c_i = k \mid y_i, x_i)$. Consequently, we estimate $Q(c_i)$ based on both the input images $x_i$ (since representations are derived from these images) and the class labels $y_i$ (since clustering is performed only on diseased samples, $y = 1$). Meanwhile, for healthy samples, the clustering pseudo-labels are assigned a uniform distribution, \textit{i.e.}, $1/K$.
    \item To maximize the lower bound (and consequently the conditional joint likelihood), the KL divergence term should be minimized. Since KL divergence is always non-negative, the optimal scenario occurs when it is zero. This is achieved when the two distributions, $Q(c_i) \approx p(c_i = k \mid y_i, x_i)$ and $p_{\theta}(c_i \mid x_i)$, are identical.
\end{itemize}
\REBUTTAL{In addition to the above differences, our method introduces the following novel elements absent in prior work: (1) Sinkhorn–Knopp regularization during clustering, which uses an entropy-regularized optimal transport penalty to balance assignments and avoid collapsed clusters; (2) a subgroup re-identification step, wherein we solve a label-assignment problem between EM iterations to maintain consistent cluster identities; and (3) a deep mixture-of-experts loss, where we train one classifier (“expert”) per cluster in a unified EM objective, enabling joint representation learning and subtype classification.}

\subsection{Expectation step \label{sec:expectation}}
During the Expectation step, we freeze the current estimate of the encoder parameters $\theta^{(t)}$, at epoch $t$, and estimate $Q^{(t)}$ as $p_{\theta^{(t)}}(c_i = k | y_i, x_i), \forall i \in |[1, n]|, \forall k \in |[1, K]| $, see Eq. \ref{eq:3}.
In order to do that, since we assume that only the positive class ($y_i=+1$) contains subgroups, we first compute $p_{\theta}(c_i = k | x_i, y_i=+1)$ using a regularized K-mean. 
%on the scaled diseased representations (with a Robust or a Standard scaler, depending on the dataset).
Please note that any clustering algorithm could be used here to compute $p_{\theta}(c_i = k | x_i, y_i=+1)$ depending on prior knowledge about the subgroups' number, size, distribution, or density.  Here, we assume that the number of subgroups $K$ is known, and thus K-means seems to be a reasonable and simple choice. 
Concerning samples from the healthy class ($y_i=-1$), as we aim to estimate sub-groups within diseased samples only, we propose to use a uniform clustering probability distribution: 
\begin{equation}
   p_{\theta}(c_i = k | y_i=-1, x_i) = \frac{1}{K} \quad  \forall i \in |[1, n]|, \forall k \in |[1, K]| 
\end{equation}
\noindent This means that healthy samples have an equal probability of belonging to the subgroups and, as detailed in Sec.~\ref{sec:maximization}, this will be used to regularize the representation so that samples from the healthy class are equidistant, in the representation space, from the subgroups centroids. By applying this strategy, the general variability, common to both healthy and diseased classes, should be disregarded in the representation space, which should be structured only by the pathological variability of the diseased class. 
%Eventually, one may strengthen this constraint by adding a weight on the ''healthy'' samples pseudo-label regularization. \PG{not clear}\\
%In practice, we found it necessary to add a weight of $10$ on this term in a difficult case, the neuro-psychiatric experiment. \\
As in Deep Cluster \cite{caron_2018}, during the subgroup re-estimation, the subgroup membership $c_i$ is re-estimated at each epoch for each sample. This entails two potential issues that have to be dealt with: subgroup degeneracy and subgroup re-identification. 

%in order to estimate $p_{\theta^{(t)}}(c_i = k|y_i,x_i)$. From Eq.~\ref{eq:3}, we then estimate $Q^{(t+1)} = p_{\theta}(c_i = k | y_i, x_i), \forall i \in |[1, n]|, \forall k \in |[1, K]| $ in order to tighten the proposed lower bound. 

%\noindent Similarly to related works, we propose to estimate $Q$ at initialization and at each epoch with K-Means algorithm on representations, as in Deep Cluster \cite{caron_deep_2018}, PCL \cite{li_prototypical_2021}. Differently, as we place our method in a Subtype Discovery paradigm, we fit and predict clustering on positive samples only.

%\noindent The pseudo-labelling process needs several tricks in order to avoid degeneracy. Inspired by Deep Clustering works, we proposed the use of Sinkhorn-Knopp regularization to avoid empty/underrepresented clusters. Additionally, we introduce another method to re-identify clusters between each epoch.

\subsubsection{Subgroup degeneracy: } In DeepCluster \cite{caron_2018}, authors have observed that K-Means may yield imbalanced or empty clusters. In order to avoid that, we use a Sinkhorn-Knopp (SK) \cite{cuturi_2013} regularization embedded into soft K-Means. Our method combines thus the clustering expressivity of soft K-Means with the equipartition regularization of the SK algorithm, as in \cite{caron_2020}. Furthermore, one can keep these soft pseudo-labels ($0 \leq Q(c)\in \mathcal{R} \leq 1$) or make them hard ($Q(c) \in \{0,1 \}$) using the OneHot encoding function.
%$Q(c)$ can be estimated using either soft pseudo-labels ($0 \leq Q(c)\in \mathcal{R} \leq 1$) or hard pseudo-labels ($Q(c) \in \{0,1 \}$). 
Similarly to \cite{sun_2021}, we propose to linearly interpolate $Q(c)$ from soft to hard probabilities along the epochs. In this way, if the initialization is not good, the use of soft pseudo-labels at the beginning of the training avoids fitting unreliable hard pseudo-labels. 
%In the case where initialization is unsatisfactory, we propose to use soft pseudo-labels at the beginning of the training in order to avoid fitting unreliable pseudo-labels. 
Instead, toward the end of the training, hard pseudo-labels are preferred to avoid under-fitting. To justify this choice, we conducted an ablation study on MNIST in Tab~\ref{table:MNIST_7_clustering_results}. In our algorithm, we first initialize the subgroups centroids $\mu = \{ \mu^k \}_{k \in |[ 1, K]| }$ with the K-Means ++ algorithm. Then, we compute the soft probabilities $Q(c_i=k) = \frac{1/||f_\theta(x_i)-\mu^k||^2_2}{\sum^K_{a=1} (1/||f_\theta(x_i)-\mu^a||^2_2)}$ only for positive samples and encourage their equipartition across the subgroups via the SK regularization: $Q' = SK(Q, \epsilon=0.05)$, as in \cite{caron_2020} (see Implementation Details in Appendix for a Discussion on $\epsilon$), where $Q' \in \mathbf{(0,1)}^{N^+ \text{x} K}$. We then interpolate between soft and hard pseudo-labels using:  $\hat{Q} = \frac{T-t}{T} \text{OneHot}(Q') + \frac{t}{T} Q' $, where $t$ is the current epoch and $T$ is the total number of epochs, and finally update the centroids. 
% At inference time, we compute the subgroup probabilities with the same formula as above.
\subsubsection{Subgroup re-identification} 
At epoch $t+1$, since we re-estimate the centroids of each subgroup, we need to identify which was the corresponding centroid at epoch $t$
%When an updated clustering is estimated, we have to identify each updated cluster (epoch $t+1$) with its most similar previous cluster (epoch $t$) 
in order not to disrupt the training process. To guarantee that, we introduce a new permutation operation $\sigma(.)$ from the labels (i.e., $1,..,K$) of the centroids at epoch $t$ to the ones at epoch $t+1$. For instance, $\sigma(1)=2$ means that the second centroid at epoch $t+1$ corresponds to the first centroid at epoch $t$. Being a permutation, we would like it to be a bijective mapping in order not to merge two subgroups into one and thus create empty subgroups (e.g., $\sigma(k) = \varnothing$). This issue could occur by naively assigning to each updated centroid (epoch $t+1$) the label of its most similar previous centroid (epoch $t$). 
%We assume that the positions of the centroids of all subgroups at epoch $t+1$ will not change much with respect to epoch $t$, in particular with respect to the other centroids. From a mathematical point of view, this means assuming that $d(\mu^{k,(t)}, \mu^{k,(t+1)}) < d(\mu^{k,(t)}, \mu^{a,(t+1)})$ $\forall k,a$.
%For example, let us assume that the $t$-epoch clustering identifies two subgroups with centroids $\mu_0^t$, $\mu_1^t$, and that the $(t+1)$-epoch clustering identifies updated subgroups with centroids $\mu_0^{t+1}$, $\mu_1^{t+1}$. 
%Ideally, assuming that 
%
%each subgroup at epoch $t+1$ is a small revised versions of the corresponding subgroup at epoch $t$ (i.e.: small centroid changes).
%, we would like to ensure a certain continuity of the subgroups labeling: $s(\mu_0^t, \mu_0^{t+1}) \geq s(\mu_0^t, \mu_1^{t+1})$, and $s(\mu_1^t, \mu_1^{t+1}) \geq s(\mu_1^t, \mu_0^{t+1})$, where $s$ is a similarity measure between the centroids.
%Concerning clustering re-identification between each epoch, we propose a method
%on the labels of freshly updated subgroups. 
To avoid that, we first compute the similarity matrix (i.e., normalized dot product) of size $K \times K$ between the $K$ previous centroids (epoch $t$) and the $K$ updated centroids (epoch $t+1$).
%The similarity between each previous and updated cluster is calculated via the normalized dot product between their centroids. 
%Naively, one could assign to an updated centroid (epoch $t+1$) the label of its most similar previous centroid (epoch $t$). 
%However, it potentially allows for more than one previous centroid to be merged into a single updated cluster (e.g.: $\sigma(0) = 1$, $\sigma(1) = 1$), which may produce one or more empty clusters (e.g.: $\sigma^{-1}(0) = \varnothing$).
%Indeed, given $K$ non-empty old clusters, if the %identifiability identification function between old and new clusters is a surjective function, new clustering will have less than $K$ non-empty clusters.
%To avoid this problem, we constraint $\sigma$ to be a bijective mapping. To do so, 
Then, we apply the optimal transport algorithm using the similarity matrix and the Sinkhorn-Knopp regularization,  so that we assign to each cluster at epoch $t+1$ the label of its most similar cluster at epoch $t$, respecting at the same time the equipartition constraint and thus the bijective mapping. Indeed, since we have $K$ centroids for $K$ labels, the equipartition property is respected if and only if each centroid at epoch $t+1$ is mapped to a single label, which is equivalent to having a bijective mapping between centroids at epoch $t$ and $t+1$.
%Namely, we apply the SK regularization to assign to each updated centroid a single former centroid's label.
% equally distribute new centroids across previous clusters in an optimal way %Therefore, the identifiability function between old and new clusters is bijective, which avoids 
% thus avoiding less-represented clusters to be merged with others. 
The implementation is straightforward and requires no extra hyperparameter tuning, see code GitHub.

\subsection{Maximization step}\label{sec:maximization}
\noindent We freeze the current estimate of $Q^{(t)} = p_{\theta^{(t)}}(c_i|x_i,y_i)$ and maximize Eq.~\ref{eq:4} to estimate the classifying experts $p_{\theta,\phi}(y_i | x_i, c_i=k)$ and the clustering head $p_{\theta,\psi}(c_i=k | x_i)$.

%optimize for $\theta$, $\phi$ and $\psi$ by maximizing Eq.~\ref{eq:5}.
%given observed data $X$ and estimated clusters $C$. 

%\begin{comment}
%    \newline Now that $Q^{(t)}$ is fixed, we can filter out the constant term $- Q^{(t)} \log Q^{(t)}$ from the maximization formula. Note that the entropy term $- Q^{(t)} \log Q^{(t)}$ is implicitly maximized during Expectation step when applying a Sinkhorn-Knopp regularization that equally re-distribute samples in clusters.
%    Therefore, Eq.~\ref{eq:5} then becomes:
%    \begin{equation}
%         \tag{6}
%         \sum_{i=1}^n  \sum_{k = 1}^K Q^{(t)} \log p_{\theta, \phi}(y_i | c_i=k, x_i) - \sum_{i=1}^n  \sum_{k = 1}^K  Q^{(t)} \log p_{\theta, \psi}(c_i=k | x_i)
%         \label{eq:6}
%    \end{equation}
%\end{comment}

\subsubsection{Mixture-of-Classifying Experts loss: }
The classifying experts $p_{\theta,\phi}(y_i | x_i, c_i=k)$ are modeled as a single neural network with $K$ outputs, one per subgroup. Let $p_{\theta,\phi_k}(y_i=+1| x_i, c_i=k)=S(h_{\phi_k}(f_{\theta}(x_i)))$ be the output prediction associated to the subgroup $k$, where $S$ is a sigmoid activation function. Equivalently, we have $p(y_i=-1 | x_i, c_i=k) = 1 - S(h_{\phi_k}(f_{\theta}(x_i)))$.  For each subgroup $k$, we can thus interpret the label $y_i$ as a Bernoulli random variable with $\text{Pr}(y_i=+1)=p_k = S(h_{\phi_k}(f_{\theta}(x_i)))$ and rewrite the first term of Eq.~\ref{eq:4} as:
$ Q{(c_i = k)} \log \big(  \frac{p_k^{y_i}}{(1-p_k)^{y_i}} \big)= Q{(c_i = k)} \big( y_i \log p_k  - y_i \log ( 1- p_k) \big) $, which is a weighted binary cross entropy between ground truth labels $y_i$ and output predictions $p_k$.  The maximization of this term is equivalent to the estimate of $K$ sub-classifiers, one per subgroup. This approach is known as a Mixture-of-Experts (MoE) \cite{tsai_2021}, \cite{zixiang_2022} where an expert $k$ is specialized in discriminating the subgroup $k$ samples from the negative class. Even if all the expert's classifiers are trained with the same objective function (i.e., a weighted binary cross-entropy), they are expected to converge towards different solutions since they are differently weighted by $Q{(c_i = k)}$. Please note that the proposed method is different from the standard MoE routing mechanism where the gating function seeks clusters \textit{across all classes} ($Q(c)=p(c|x)$) \cite{zixiang_2022}, whereas our gating function seeks subgroups \textit{only within the positive class} ($Q(c)=p(c|x,y)$). % This brings a different mathematical formulation and optimization procedure.

\subsubsection{Clustering head loss: } %\label{clustering}
%As activation function in the output layer, we use a sigmoid and a softmax for classifying experts and classification head respectively.
The clustering head $p_{\theta, \psi}(c_i | x_i)=\sigma(h_{\psi}(f_\theta(x_i)))$ is modeled as a neural network with a softmax function $\sigma$ as output activation function since it predicts cluster probabilities. The parameters $\theta$ of the encoder and the parameters $\psi$ of the clustering network are updated through the maximization of the second term of Eq.~\ref{eq:4} (i.e., clustering regularization term). It represents the Kullback-Leiber divergence between the subgroup pseudo-labels $Q(c_i)$ (estimated to approximate $p_{\theta}(c_i | y_i, x_i)$) and the clustering head predictions $p_{\theta, \psi}(c_i | x_i)$.  This loss aims at minimizing the discrepancy between the two distributions $p_{\theta}(C | Y, X)$ and $p_{\theta, \psi}(C | X)$, 
%During the Expectation step, we estimated $Q$ using a regularized K-means for the disease class samples ($y_i=+1$) and a uniform distribution (i.e., $1/K$) for healthy samples ($y_i=-1$). 
%Forcing the unsupervised clustering prediction $p_{\theta, \psi}(c_i | x_i)$ to be equal to $Q(c_i)$ 
producing a representation space more suited for subgroup discovery. Indeed, negative samples should be encoded in the representation space as points equidistant from the subgroup centroids since their membership probability should be the same for all subgroups (i.e., $1/K$). Furthermore, positive samples should be clustered as in $Q(c_i)$, namely as if the ''unsupervised'' clustering algorithm was only considering the pathological/positive variations. This regularization promotes a representation space where the general variability (common to both negative and positive classes) is discarded for the identification of subgroups. \\

\noindent \REBUTTAL{We further analyze the effect of this constraint by connecting it to the alternatives suggested in the literature. Assigning controls to a dedicated cluster (as in WS-DeepClustering) isolates them but does not force the encoder to disentangle shared from disease-specific variability. Promoting global uniformity over the representation space (as in SimCLR and SupCon) regularizes all samples indiscriminately. In contrast, our uniform cluster assignment selectively targets controls: it enforces high entropy in the clustering head for controls while preserving structured, low-entropy assignments for disease samples. This shapes the latent geometry so that subgroup centroids lie at comparable distances from the control region.} Using the MNIST example, we plot a 3D visualization of the representation space in Appendix, Fig. \ref{fig:umap_evolution}. We observe that both bold and thin digits of the negative class (i.e., all digits but the 7) are encouraged to be equidistant from the subgroups of the digit 7 in the representation space across the epochs (MNIST: distance ratio from 3.68 to 1.35 during training, Appendix~E). \REBUTTAL{Combining this constraint with Sinkhorn–Knopp regularization proves critical, as Deep UCSL with Sinkhorn–Knopp consistently outperforms its counterpart without it.}

\textbf{Mixture-of-experts: }
At inference, one can compute the classification label $y_j$ for a new test sample $x_j$ using the Mixture-of-Experts prediction, defined as:
\begin{equation}
    p(y_j | x_j) = \sum_{k = 1}^K p_{\theta, \phi_k}(y_j | x_j, c_j=k) p_{\theta, \psi}(c_j=k | x_j)
\end{equation}

%%%%%%%%%%%%%%%%%%%%%%%%%%%%%%%%%%%%%%%%%%%%%%%%%%%%%%%%%%%%%%%%%%%%%%%%%%%%%%%%%%%
\section{Experiments}
\noindent This section evaluates Deep UCSL and compares it with several SOTA (State-Of-The-Art) methods on a synthetic dataset (Digit 7 MNIST) (with an Nvidia K80), three medical image applications (with an NVIDIA V100) and a neuro-psychiatric application (with an NVIDIA RTX3080). Results uncertainties (\textit{i.e.,} $\pm$) are obtained with $3$ different initializations evaluated on the same independent test set. Only for the neuro-psychiatric case, the variability is instead obtained from 5 different TRAIN/VAL splits (0.9, 0.1) whose models are evaluated on the same external TEST set.

\subsection{Implementations and evaluation criteria.}
We train all methods using only the class label $y$ (healthy versus disease), but \textbf{not} the subgroup labels $c$. Then, to quantitatively evaluate performance, we use TEST sets where we know both the class label $y$ and the subgroup label $c$. All hyperparameters used in the experiments are available in the appendices or in the open-source code provided. For methods incorporating a supervised classification prediction head (Cross-Entropy, Deep UCSL, SupCon), the models were trained to classify samples as either diseased or healthy. The learned representations of the training samples were then clustered using k-means to infer pathological subgroups. For unsupervised methods (DeepCluster, PCL, SimCLR, BYOL, SCAN), models were trained only on diseased samples. Similarly, the learned representations were clustered using k-means to identify pathological subgroups. Since the representation/contrastive learning methods do not have a classification head (DeepCluster, PCL, SimCLR, BYOL, SCAN), we test their performance only in subgroups identification with a K-means algorithm (the same as in our method) fitted only on target samples. This means that these methods are evaluated only on the subgroup discovery part, as if they had a perfect classification head (no diagnosis error), thus ensuring a fair comparison. Additional experiments led with a weak-supervised deep clustering method WS-DeepClustering as well as domain-specific (ophtalmology: RetFound \cite{zhou_2023} / radiology: EVA-x  \cite{yao_2024}) pre-trained foundation models were also assessed. \REBUTTAL{WS-DeepClustering is a weakly supervised adaptation of DeepCluster where clustering is performed via alternating k-means assignments (trained on diseased samples only) and encoder updates between predictions and pseudo-labels (using a cross-entropy objective with $K+1$ classes), with control samples assigned to a dedicated cluster (the $K+1$th)}. More experimental details are described in Appendix's Additional Results. The evaluation metrics were the following:
\begin{itemize}
    \item \textbf{Class Balanced Accuracy (Class B-ACC):} Measures the balanced accuracy of binary classification between true labels (healthy: 0, disease: 1) and predicted diagnoses $p(y_j | x_j)$.
    \item \textbf{Subgroup Balanced Accuracy (Subgroup B-ACC):} Evaluates the balanced accuracy between inferred pathological subgroups compared to the true subgroups $c_j$ (not used during training, only during evaluation). This metric is thus computed on disease samples only.
    \item \textbf{Overall Balanced Accuracy (Overall B-ACC):} Accounts for both classification and subgroup prediction errors: $\frac{1}{2} \frac{TP}{TP + FN} + \frac{1}{2} \frac{TN}{TN + FP}$, where  $TN$ and $FN$ are the class true and false negatives, namely the number of healthy and disease samples classified as healthy, respectively. $TP$ is the number of disease samples correctly classified \textit{AND} assigned to the right subgroup. $FP$ is the number of healthy samples classified as disease \textit{OR} disease samples correctly classified but assigned to the wrong subgroup. This metric reflects both diagnosis errors (e.g., misclassifying a healthy sample as diseased) and subgroup misclassification errors (e.g., assigning a sample to the wrong pathological subgroup, such as viral instead of bacterial pneumonia).
\end{itemize}

\subsection{Digit MNIST dataset}
\begin{figure}[!ht]
    \centering-
    \includegraphics[width=.8\linewidth]{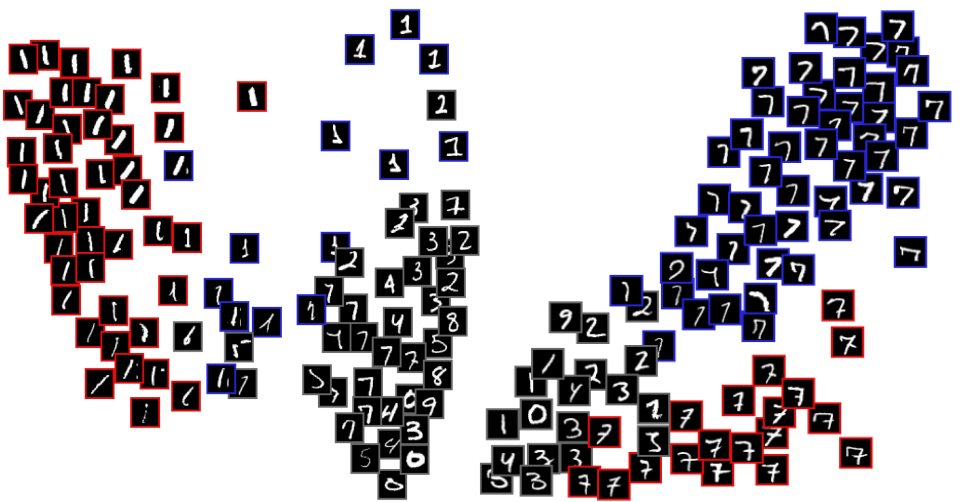}
    \caption{MNIST qualitative results. 2D PCA plots of the representation spaces learned by Deep UCSL when seeking two subgroups for digits 1 (left) and 7 (right).
    PCA is trained on the TRAIN set and then used on the TEST set for visualization purposes. }
    \label{fig:mnist_digits_sutyping_examples}
\end{figure}

We constructed two MNIST-based datasets for subgroup discovery by treating digits 7 and 1 as the pathological class and the remaining digits as healthy. Each positive digit admits two visually meaningful subgroups: 7s with or without a middle crossbar, and 1s with or without an upper diagonal stroke (cf. Fig.~\ref{fig:mnist_digits_sutyping_examples}). Despite strong class imbalance (7: 20–80\%, 1: 5–95\%), Deep UCSL successfully uncovers these latent subgroups. For digit 7, we manually labeled 400 test samples to evaluate subgroup identification. Deep UCSL outperforms unsupervised clustering, representation learning, and supervised baselines (Table~\ref{table:MNIST_7_clustering_results}). Unlike contrastive learning approaches, whose performance relies on carefully designed augmentations to suppress irrelevant variability, Deep UCSL automatically discards variability shared between healthy and diseased samples and focuses on pathology-specific patterns. Finally, compared to the linear UCSL \cite{louiset_2021}, which operates directly on pixels and is sensitive to spatial shifts, Deep UCSL leverages convolutional feature extractors to learn more robust, nonlinear representations.
\begin{table}[ht!]
    \vspace{-20pt}
    \caption{MNIST experiment about subgroup discovery of digit 7.
    Top (unsupervised) methods are trained on 7 digits only. ''\textit{Morpho}'' methods have morphological data augmentations that simulate digit boldness. For Representation Learning methods, clusters were fitted and inferred using K-Means on the representations of positive samples only. The Class Accuracy (7 vs rest) is always 100\%. (*) indicates that Deep UCSL is significantly better ($p \leq 0.05$) than other methods using a two sample t-test.}
    \setlength\tabcolsep{0.12\textwidth}
    \centering
    {\footnotesize
    \begin{tabular}{cccccccc} 
    \hline
    Algorithm & Subgroup B-ACC \\
    \hline
    Deep Cluster-v2 \cite{caron_2018}, \cite{caron_2020} & 0.540$\pm$0.032 \\  
    PCL \cite{li_2021} & 0.555$\pm$0.018 \\
    \textit{Morpho} PCL & 0.732$\pm$0.027 \\
    BYOL \cite{grill_2021} & 0.552$\pm$0.013 \\
    SCAN \cite{van_gansbeke_2021} & 0.597$\pm$0.039\\
    SwAV \cite{caron_2020} & 0.601$\pm$0.009 \\
    SimCLR \cite{chen_simclr_2020} & 0.634$\pm$0.007 \\ 
    \textit{Morpho} SimCLR & 0.721$\pm$0.021 \\
    AE + K-Means & 0.776$\pm$0.007 \\
    \hline
    MoE \cite{zixiang_2022} & 0.5$\pm$0.0 \\
    BCE + K-Means & 0.633$\pm$0.020 \\
    SupCon \cite{khosla_2020} & 0.669$\pm$0.066 \\ 
    \textit{Morpho} SupCon & 0.808$\pm$0.018 \\ 
    UCSL \cite{louiset_2021} & 0.815$\pm$0.009 \\
    \hline
    WS-DeepClustering & 0.744$\pm$0.019 \\
    Deep UCSL with soft pseudo-labels & 0.816$\pm$0.017 \\
    Deep UCSL with hard pseudo-labels & 0.895$\pm$0.014 \\
    Deep UCSL without SK & 0.908$\pm$0.008 \\ 
    Deep UCSL & \bf{0.920$\pm$0.015}* \\ 
    \hline
    \end{tabular}
    }
    \vspace{-20pt}
    \label{table:MNIST_7_clustering_results}
\end{table}

\subsection{Neuro-psychiatry application.} 
\noindent We create another dataset for subgroup identification comprising 3D MRI T1 weighted images of the brain. 
The healthy class contains Healthy Controls (HC=686), and the disease class, Mental Disorders (MD), comprises two subgroups: 
1) patients with Schizophrenia (SZ=275), from SCHIZCONNECT \cite{wang_2016}, and 
2) patients with Bipolar Disorder (BD=307), from BIOBD dataset \cite{sarrazin_2018}. 

\noindent Please note that this is a challenging subgroup discovery problem given the: high dimensionality of 3D brain images, the subtle differences between healthy controls and patients, the few training data, the continuum between both diseases, and the different acquisition machines/protocols \cite{dufumier_exploring_2024}. In Table \ref{table:BP_vs_SZ_results}, we show the subgroup identification capability of Deep UCSL compared with related works.  All evaluation criteria are computed on an independent TEST set (199 HC, 190 SZ, 116 BP), coming from the BSNIP cohort \cite{tamminga_2014}, with different acquisition sites.  Controls and patients share common (thus irrelevant) sources of variations (e.g.: age, sex, acquisition site).

\noindent Interestingly, UCSL performance is highly sensitive to the choice of input features: it degrades when using generic VAE latent representations but improves with carefully engineered domain-specific features such as age-corrected Gray Matter Region Based-Morphometry. In contrast, Deep UCSL offers an end-to-end approach that removes the need for prior feature engineering while achieving comparable or better performance.
\vspace{-20pt}
\begin{table}[ht!]
    \caption{Results on Neuro-psychiatry task (BP/SZ) on an independent TEST set. top methods are trained on [SZ+BP] only. (*) indicates that Deep UCSL is statistically significant better ($p \leq 0.05$) than all other methods using a two sample t-test. % "CE + K-Means" is a binary classification method on which we applied a K-Means on the representation. 
    %All results were obtained with K-Means predictions on the target class representations (rep), or the projection heads (head).
    }
    \centering
    \setlength\tabcolsep{0.035\textwidth}
    {\footnotesize
    \begin{tabular}{cccccccc} 
    \hline
    Algorithm & Subgroup B-ACC & Class B-ACC & Overall B-ACC \\
    \hline
    Deep Cluster - v2 \cite{caron_2018}, \cite{caron_2020} & 0.517$\pm$0.010 & $\times$ & $\times$ \\ 
    PCL \cite{li_2021} & 0.542$\pm$0.030 & $\times$ & $\times$ \\
    SwAV \cite{caron_2020} & 0.522$\pm$0.008 & $\times$ & $\times$ \\
    SCAN \cite{van_gansbeke_2021} & 0.509$\pm$0.008 & $\times$ & $\times$ \\
    SimCLR \cite{chen_simclr_2020} & 0.571$\pm$0.017 & $\times$ & $\times$ \\ % head
    BYOL \cite{grill_2021} & 0.508$\pm$0.006 & $\times$ & $\times$ \\ % head
    \hline
    VAE \cite{kingma_2014} + UCSL \cite{louiset_2021} & 0.534$\pm$0.016 & 0.588$\pm$0.013 & 0.459$\pm$0.018 \\ BCE + K-Means & 0.507$\pm$0.005 & 0.653$\pm$0.025 & 0.428$\pm$0.038 \\
    SupCon \cite{khosla_2020} & 0.550$\pm$0.014 & 0.656$\pm$0.017 & 0.458$\pm$0.017 \\ % head
    GM ROI features \cite{gaser_2022} + UCSL & \textbf{0.590$\pm$0.016} & 0.653$\pm$0.012 & 0.525$\pm$0.011 \\
    Deep UCSL & \textbf{0.589$\pm$0.011} & \textbf{0.671$\pm$0.018} & \textbf{0.543$\pm$0.014}* \\ 
    \hline
    CE (upper bound) & 0.615$\pm$0.007 & $\times$ & $\times$ \\
    \hline
    \end{tabular}
    }
    \label{table:BP_vs_SZ_results}
\end{table}
\vspace{-30pt}

\subsection{Pneumonia subgroup identification.}
\begin{figure*}[!tbp]
    \vspace{-20pt}
    \centering
    % -------------------- Pneumonia --------------------
    \begin{subfigure}[t]{0.41\textwidth}
        \centering
        \includegraphics[width=0.49\linewidth]{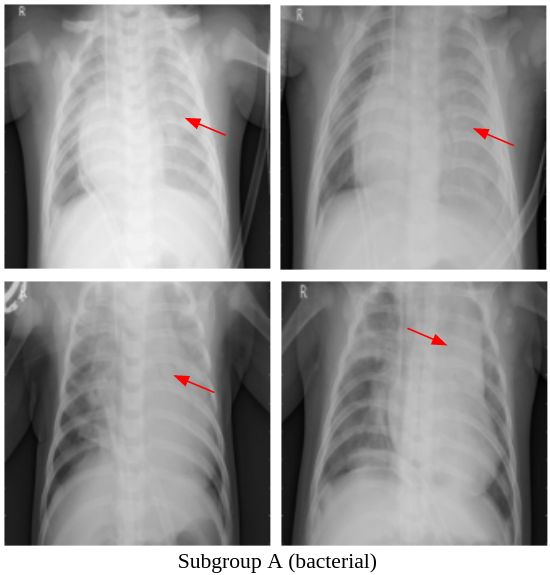}
        \includegraphics[width=0.49\linewidth]{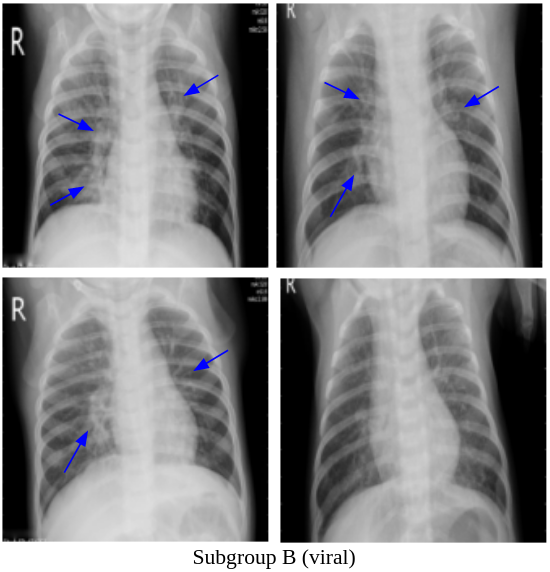}
        \caption{Bacterial pneumonia (left) shows lobar consolidation (red), whereas viral pneumonia (right) exhibits diffuse interstitial patterns (blue) \cite{kermany_2018}.}
        \label{fig:pneumonia_subtyping_examples}
    \end{subfigure}
    \hfill
    % -------------------- Retina --------------------
    \begin{subfigure}[t]{0.57\textwidth}
        \centering
        \includegraphics[width=\linewidth]{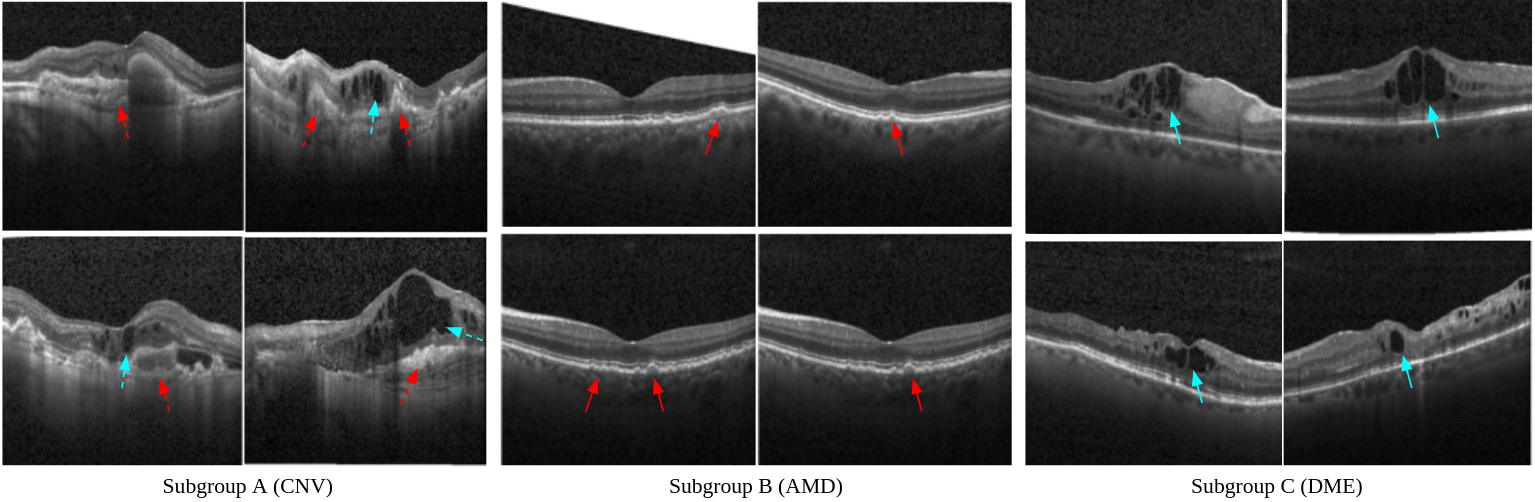}
        \caption{Cluster A: choroidal neovascularization (CNV); Cluster B: drusens in early AMD; Cluster C: diabetic macular edema (DME), with characteristic pathological patterns exhibited.}
        \label{fig:retinal_qualitative_results}
    \end{subfigure}

    \caption{Examples of images with high subgroup predictions $p_{\theta,\phi}(c \mid x)$ across two medical imaging modalities. Left: pneumonia subtypes. Right: retinopathy subtypes.}
    \label{fig:qualitative_subgroup_results}
\end{figure*}

\noindent Here, we propose to address the identification of two subgroups in pediatric medicine: viral pneumonia and bacterial pneumonia. 
We use the same training and testing datasets as in \cite{kermany_2018}. 
The testing set contains $234$ healthy samples, $242$ bacterial samples, and $148$ viral samples. We present quantitative results in Table~\ref{table:pneumonia_subtype_discovery_results}, where Deep UCSL is the best performing method in the subgroup identification task. 
In Fig.~\ref{fig:pneumonia_subtyping_examples}, we display the nearest images from each subgroup centroid. 
We observe distinct pathological patterns that are specific to bacterial and viral pneumonia. 
These satisfactory results illustrate how practitioners can leverage Subgroup Discovery to stratify a pathology.

\begin{table}[ht!]
    \caption{Comparison of different methods for the viral/bacterial subgroup identification along with diagnosis classification. (*) indicates that Deep UCSL is statistically significant better ($p \leq 0.05$) than all other methods using a two sample t-test. Upper methods are trained on disease samples only.}
    \centering
    \setlength\tabcolsep{0.045\textwidth}
    {\footnotesize
    \begin{tabular}{cccccccc} 
    \hline
    Algorithm & Class B-ACC & Subgroup B-ACC & Overall B-ACC  \\
    \hline
    DeepCluster-v2 \cite{caron_2018}, \cite{caron_2020} & $\times$ & 0.814$\pm$0.008 & $\times$ \\
    PCL \cite{li_2021} & $\times$ & 0.773$\pm$0.055 & $\times$ \\
    SwAV \cite{caron_2020} & $\times$ & 0.815$\pm$0.006 & $\times$ \\
    SCAN \cite{van_gansbeke_2021} & $\times$ & 0.576$\pm$0.059 & $\times$ \\
    SimCLR \cite{chen_simclr_2020} & $\times$ & 0.741$\pm$0.027 & $\times$ \\
    BYOL \cite{grill_2021} & $\times$ & 0.748$\pm$0.034 & $\times$ \\
    \hline
    VAE \cite{kingma_2014} + UCSL \cite{louiset_2021} & 0.734$\pm$0.020 & 0.731$\pm$0.004 & 0.646$\pm$0.007 \\
    BCE + K-Means & \textbf{0.917$\pm$0.012} & 0.560$\pm$0.014 & 0.752$\pm$0.007 \\ 
    SupCon \cite{khosla_2020} & 0.895$\pm$0.004 & 0.576$\pm$0.036 & 0.744$\pm$0.008 \\
    WS-DeepClustering & 0.881$\pm$0.008 & 0.826$\pm$0.004 &  0.812$\pm$0.007 \\
    EVA-X \cite{yao_2024} & 0.729 & 0.813 & 0.674 \\
    Deep UCSL without SK & 0.880$\pm$0.019 & \textbf{0.847$\pm$0.037}* & \textbf{0.812$\pm$0.017}* \\
    Deep UCSL & 0.886$\pm$0.010 & \textbf{0.835$\pm$0.007}* & \textbf{0.820$\pm$0.012}* \\
    \hline
    CE (upper bound) & $\times$ & 0.891$\pm$0.005& $\times$ \\
    \hline
    \end{tabular}
    }
    \label{table:pneumonia_subtype_discovery_results}
\end{table}

\subsection{Retinal pathology applications}
\noindent We ultimately validate our method on the discovery of disease subgroups in retinal pathologies. 
We evaluate our performance on two different datasets. First, we use the Retinal Pathology OCT (Optical Coherence Tomography) dataset introduced in \cite{kermany_2018}.  The test set comprises $242$ healthy samples and $242$ samples for each pathological subgroup. Quantitative results are shown in Table~\ref{table:subtype_discovery_results_OCT}.  Deep UCSL is placed among the best performing methods in terms of binary classification ("Class B-ACC") and Subgroup Discovery ("Subgroup B-ACC").  In terms of "Overall B-ACC", SupCon closes the gap with Deep UCSL, but Deep UCSL remains statistically significantly better.  As in the previous section, qualitative results in Fig.~\ref{fig:retinal_qualitative_results} show that the closest TEST images to subgroup centroids exhibit distinct pathological patterns specific to each retinal pathology.  Again, these satisfactory results illustrate how practitioners can leverage a Subtype Discovery method to stratify and better interpret pathological images. We also use the Ocular Disease Intelligent Recognition (ODIR) dataset\footnote{\url{https://odir2019.grand-challenge.org/}}, which contains 1,890 healthy and 1,391 diseased patients divided heterogeneously into five disease subgroups: Diabetes, Glaucoma, Cataracts, Age-related macular degeneration, and pathological Myopia.  The test set contains $210$ healthy and $155$ pathological samples, divided heterogeneously across disease subgroups.   Table~\ref{table:subtype_discovery_results_ODIR} displays quantitative results.  Deep UCSL performs better than all other state-of-the-art methods on Subgroup B-ACC and Overall B-ACC.
\begin{table*}[ht!]
\centering
\setlength\tabcolsep{0.001\textwidth}

\begin{minipage}{0.51\textwidth}
\centering
\caption{RetOCT\cite{kermany_2018} (3 subgroups). (*): Deep UCSL is significantly better ($p\leq 0.05$). Upper methods trained on diseased samples only.}
{\footnotesize
\begin{tabular}{lccc}
\hline
Algorithm & Class & Subgroup & Overall \\
 & B-ACC & B-ACC & B-ACC \\
\hline
DeepCluster-v2 & $\times$ & 0.593$\pm$0.052 & $\times$ \\
PCL & $\times$ & 0.428$\pm$0.022 & $\times$ \\
SwAV & $\times$ & 0.563$\pm$0.073 & $\times$ \\
SCAN & $\times$ & 0.449$\pm$0.045 & $\times$ \\ 
SimCLR & $\times$ & 0.578$\pm$0.009 & $\times$ \\
BYOL & $\times$ & 0.337$\pm$0.002 & $\times$ \\
\hline
VAE+UCSL & 0.350$\pm$0.002 & 0.585$\pm$0.026 & 0.369$\pm$0.016 \\
BCE+K-Means & 0.996$\pm$0.003 & 0.371$\pm$0.009 & 0.672$\pm$0.001 \\
SupCon+K-Means & 0.999$\pm$0.001 & 0.618$\pm$0.002 & 0.732$\pm$0.001 \\
WS-DeepClustering & 0.993$\pm$0.002 & 0.554$\pm$0.060 & 0.705$\pm$0.018 \\
RETFound(OCT/F) & 0.9676 & 0.3650 & 0.6239 \\
Deep UCSL w/o SK & 0.999$\pm$0.001 & 0.345$\pm$0.003 & 0.668$\pm$0.001 \\
\textbf{Deep UCSL} & \textbf{1.0$\pm$0.0} & \textbf{0.626$\pm$0.010} & \textbf{0.735$\pm$0.003}* \\
\hline
CE (upper bound) & $\times$ & 0.971$\pm$0.004 & $\times$ \\
\hline
\end{tabular}
}
\label{table:subtype_discovery_results_OCT}
\end{minipage}
\hfill
\begin{minipage}{0.41\textwidth}
\centering
\caption{ODIR$^1$ (5 subgroups). (*): Deep UCSL is significantly better ($p\leq 0.05$). Upper methods trained on diseased samples only.}
\centering
{\footnotesize
\begin{tabular}{lccc}
\hline
 Class & Subgroup & Overall \\
 B-ACC & B-ACC & B-ACC \\
\hline
 $\times$ & 0.533$\pm$0.023 & $\times$ \\
 $\times$ & 0.308$\pm$0.009 & $\times$ \\
 $\times$ & 0.415$\pm$0.026 & $\times$ \\
 $\times$ & 0.467$\pm$0.043 & $\times$ \\
 $\times$ & 0.452$\pm$0.033 & $\times$ \\
 $\times$ & 0.287$\pm$0.026 & $\times$ \\
\hline
 0.532$\pm$0.030 & 0.309$\pm$0.047 & 0.469$\pm$0.023 \\
 0.728$\pm$0.006 & 0.424$\pm$0.038 & 0.548$\pm$0.027 \\
 0.716$\pm$0.002 & 0.524$\pm$0.021 & 0.575$\pm$0.012 \\
 0.714$\pm$0.014 & 0.517$\pm$0.032 & 0.581$\pm$0.028 \\
 \textbf{0.7331} & 0.3014 & 0.4664 \\
 \textbf{0.736$\pm$0.001} & 0.371$\pm$0.061 & 0.522$\pm$0.024 \\
 \textbf{0.732$\pm$0.003} & \textbf{0.560$\pm$0.020}* & \textbf{0.619$\pm$0.006}* \\
\hline
 $\times$ & 0.737$\pm$0.003 & $\times$ \\
\hline
\end{tabular}
}
\label{table:subtype_discovery_results_ODIR}
\end{minipage}

\end{table*}

\section{Conclusion}
In this work, we proposed, to the best of our knowledge, the first deep-learning method for Disease Subgroup Discovery that contrasts with healthy controls. Our work is motivated by the failure of linear methods, such as UCSL and HYDRA, when latent subgroups stem from non-linear patterns in input images, which are difficult to capture with manual feature engineering. To do so, we took inspiration from UCSL to derive our objective and motivate the use of a deep encoder network as a feature extractor. As in UCSL, we use an Expectation-Maximization optimization process to alternate between the subgroup pseudo-labels estimation and the classification of each subgroup from the healthy class. Differently from UCSL, we motivate the need for a clustering regularization to update the encoder's representation so that we can correctly discriminate the pathological subgroups and encourages healthy samples not to belong to a pathological subgroup. Furthermore, this regularization guarantees the monotonical convergence of the optimization procedure. \RL{Integrating deep neural networks into UCSL is non-trivial due to well-known deep clustering issues, namely clustering degeneracy and cluster re-identification across training epochs. Inspired by the Sinkhorn–Knopp optimal transport algorithm, we introduce two strategies to address these challenges: a bijective matching between successive cluster centroids to ensure consistent cluster identities, and a regularized Soft K-Means enforcing approximate equipartition of samples across subgroups. In many medical imaging applications, diagnosis labels are readily available while disease subgroups remain unknown. We address this by discovering disease subgroups using only binary diagnosis supervision, and demonstrate through quantitative and qualitative evaluations on controlled and real-world datasets that our method effectively identifies meaningful pathological subgroups.}

\section*{Declarations}

\noindent \textbf{Funding}  
This work was supported by the French government’s \textit{Investissements d’Avenir} programme (ANR-18-RHUS-0014 PsyCARE), the Big2Small Chair in Artificial Intelligence funded by the French National Research Agency, and the R-LiNK Horizon 2020 European Union grant (H2020-SC1-2017, 754907).

\noindent \textbf{Competing interests}  
The authors declare that they have no competing interests.

\noindent \textbf{Ethics and consent.}  
This study uses previously published and publicly available datasets. All data were collected in accordance with the ethical standards of the respective institutional and national research committees and with the Declaration of Helsinki. Ethics approval and informed consent were obtained by the original providers.

\noindent \textbf{Data availability}  
The datasets analyzed during the current study are publicly available from the original sources cited in the manuscript. 

\noindent \textbf{Code availability}  
The code used to implement Deep UCSL and reproduce the experiments is publicly available at  
\url{https://github.com/rlouiset/deep_ucsl}.

\noindent \textbf{Author contributions}  
R.L. conceived the study, developed the methodology, implemented the models, and performed the experiments.  
E.D., B.D., A.G., and P.G. contributed to the methodological design, interpretation of the results, and critical revision of the manuscript.  
All authors read and approved the final manuscript.


\bibliography{ref-clean}% common bib file

\newpage
\begin{appendices}

\section{Convergence guarantee}
\label{convergence}
Here, we provide proof that the proposed Expectation-Maximization optimization process yields a monotonic increase of the log of the joint conditional likelihood $F(\theta,\phi,\psi)$, that is our cost function:

\begin{equation}
    \begin{aligned}
         F(\theta,\phi,\psi) &= \sum_{i=1}^n \log \left( \sum_{k=1}^K  Q(c_i=k) \frac{p_{\theta,\phi,\psi}(y_i, c_i=k | x_i)}{Q(c_i=k)} \right) \\ 
         & \geq \sum_{i=1}^n  \sum_{k=1}^K Q(c_i=k) \log p_{\theta,\phi}(y_i | c_i=k, x_i)  \\ 
         & - D_{KL}(Q(c)||p_{\theta,\psi}(c | x))
    \end{aligned}
    \label{eq:5}
\end{equation}

Given an estimation of the parameters $\theta^{(t)}$ at the t-th step, the E-step consists in choosing $Q^{(t)} = p_{\theta^{(t)}}(c_i | x_i, y_i)$, which makes the previous bound tight (\textit{i.e.,} equality).
% \begin{equation}
%     \begin{aligned}
%     & F(\theta^{(t)},\phi^{(t)},\psi^{(t)}) = \\ 
%     &\sum_{i=1}^n  \sum_{k=1}^K Q^{(t)}(c_i=k) \log p_{\theta^{(t)},\phi^{(t)}}(y_i | c_i=k, x_i) \\ 
%     & - D_{KL}(Q(c)||p_{\theta^{(t)},\psi^{(t)}}(c | x))
%     \end{aligned}
% \end{equation}
At the t-th M-step, we freeze $Q^{(t)}$ and we obtain the parameters $\theta^{(t+1)}$, $\psi^{(t+1)}$ and $\phi^{(t+1)}$ by maximizing the right-hand side of the equation above. Thus, obtaining:

\begin{equation}
    \tag{10}
    \begin{aligned}
     F(\theta^{(t+1)}, & \phi^{(t+1)},\psi^{(t+1)}) \geq \\ 
         & \sum_{i=1}^n  \sum_{k=1}^K Q^{(t)}(c_i=k) \log p_{\theta^{(t+1)},\phi^{(t+1)}}(y_i | c_i=k, x_i) \\ 
         & - D_{KL}(Q^{(t)}||p_{\theta^{(t+1)},\psi^{(t+1)}}(c | x)) \\
     & \geq \sum_{i=1}^n  \sum_{k=1}^K Q^{(t)}(c_i=k) \log p_{\theta^{(t)},\phi^{(t)}}(y_i | c_i, x_i) \\ 
         & - D_{KL}(Q^{(t)}||p_{\theta^{(t)},\psi^{(t)}}(c | x)) \\ 
         & = F(\theta^{(t)},\phi^{(t)},\psi^{(t)}) 
    \end{aligned}
\end{equation}

where the first inequality comes from Eq.~\ref{eq:5} and the second one is true since we look for the parameters $\theta^{(t+1)},\phi^{(t+1)},\psi^{(t+1)}$ that maximizes $F(\theta^{(t)},\phi^{(t)},\psi^{(t)})$. The above result shows that $F(\theta,\phi,\psi)$ monotonically increases.

\section{Clustering re-identification}
In the clustering re-identification paragraph, we aim to identify each updated cluster (epoch $t+1$) with its most similar previous cluster (epoch $t$). Let us clarify the notation. At epoch $t$, we have estimated $K$ subtypes, we can compute their respective centroids with the following formula: $\mu^t_k = \sum_{i=1}^N 1_{c^t_i=k} f_\theta(x_i)$.
The objective of the clustering re-identification algorithm is to permute the labels of the clusters (and their centroids) estimated at epoch $t+1$ so that there is a continuity between clusters estimated at epoch $t$ and those estimated at epoch $t+1$. In practice, we aim to compute a permutation function $\sigma$ that maps an updated cluster (epoch $t+1$) onto its most similar former cluster (epoch $t$). Given a similarity function $s(\mu, \mu')$ between two centroids $\mu$ and $\mu'$. We are seeking the optimal permutation $\sigma^*$, which maximizes the average similarity: $ \sigma^* = \max_\sigma \sum_{k = 1}^K s(\mu^t_k, \mu^{t+1}_{\sigma^{-1}(k)})$. \\
Importantly, we wish to construct a function $\sigma$ that is bijective. Indeed, a non-bijective mapping could potentially allow for more than one previous cluster to be merged into a single updated cluster, which may produce one or several empty clusters. For example, assuming that $K=2$ and that the estimated mapping gives $\sigma(0) = 1$, $\sigma(1) = 1$, then after having permuted the indices of the updated clusters, we would get $C^{t+1}_0 = \varnothing$ because $\sigma^{-1}(0) = \varnothing$). Thus, to ensure the bijectivity of $\sigma$, we propose to cast our problem into a conceptually different one. Let us explain it in detail.\\
Let assume that we are given $K$ data-points: $\{ c^{t+1}_j, j \in |[1, K ]| \}$ (in our experiment, it corresponds to the $K$ centroids of clusters estimated at epoch $t+1$). Now, assuming that we are given $K$ categories (which, in our case, correspond to the $K$ clusters estimated at epoch $t$). Given a similarity measure, the probability of a sample $j$ to belong to a given category $i$ can be computed with the following formula:
\begin{equation}
    p(c^t_i|\mu^{t+1}_j) = \frac{s(\mu^{t+1}_j, \mu^t_i)}{\sum_{k = 1}^K s(\mu^{t+1}_j, \mu^t_k)}
\end{equation}

\section{Sinkhorn-Knopp Soft K-Means}
\noindent In this section, we describe the pseudo-code algorithm (See Alg.~\ref{alg:softskkmeans}.) for the Soft K-Means algorithm regularized with Sinkhorn-Knopp \cite{cuturi_2013}. The OneHot(.) function consists of transforming a smooth probability clustering vector (e.g.: $[0.2, 0.1, 0.7]$) into the hard version (i.e.: $[0, 0, 1]$).

\newpage

\begin{algorithm}
\begin{algorithmic}[1]
\label{alg:softskkmeans}
\State  \textbf{Input:} 
\State \hspace*{\algorithmicindent} Disease representations: $Z \in \mathbf{R}^{N_{y=1} \times D}$,
\State \hspace*{\algorithmicindent} $K$: subgroups number, $\lambda$: SK temperature \State \hspace*{\algorithmicindent} $N$: iterations
\State \textbf{Output:}
\State \hspace*{\algorithmicindent} Centroids: $\mu = \{ \mu^k \}_{k \in |[ 1, K]| }$.
\State {\bf Initialization step}:
\State \hspace*{\algorithmicindent} Initialize centroids $\mu$ with K-Means ++ algorithm.
\For {i in N iterations}
    \State \hspace*{\algorithmicindent} Compute soft clustering probabilities $Q(c_i)$ given a representation $Z_i$: $Q(c_i) = \frac{1/||Z_i-\mu_i||^2_2}{\sum^K_{j=1} (1/||Z_i-\mu_j||^2_2)}$.
    \State \hspace*{\algorithmicindent} Apply SK regularization: $Q = SK(Q, \lambda)$.
    \State \hspace*{\algorithmicindent} Compute one-hot clustering matrix $Q_{hot}$:\\ \hspace*{\algorithmicindent} \hspace*{\algorithmicindent} \hspace*{\algorithmicindent} $Q_{hot} = OneHot(Q.argmax(dim=1))$.
    \For {k in K subgroups}
        \State \hspace*{\algorithmicindent} Update centroid $k$: $\mu^k = \frac{(Z * Q_{hot}[:, k]).sum()}{Q_{hot}[:, k].sum()}$.
    \EndFor
\EndFor
\State Return centroids $\mu = \{ \mu^k \}_{k \in |[ 1, K]| }$.
\end{algorithmic}
\caption{SK regularized Soft K-Means pseudo-code}
\end{algorithm} 

\newpage

\section{Pseudo-Code of Deep UCSL}
\begin{algorithm}
\label{alg:Deep_UCSL_algorithm}
\begin{algorithmic}[1]
\State  \textbf{Input:} $X \in \mathbf{R}^{N \text{x} (C*W*H)} $, $y \in \{-1, 1\}^N$, $K$: \# subgroups, $\epsilon$: SK temperature, T: \# epochs
\State \textbf{Output:}
\State \hspace*{\algorithmicindent} Features encoder: $f_\theta$
\State \hspace*{\algorithmicindent} Clustering head: $p_{\theta, \psi}(c_i | x_i)$
\State \hspace*{\algorithmicindent} Classifying Experts: $p_{\theta,\phi}(y_i | x_i, c_i=k), \forall k$
\State \hspace*{\algorithmicindent} Fitted K-Means: $p_{\theta}(c_i |  x_i, y_i)$
\State {\bf Initialization step}: Estimate $Q^{(0)}$:
\State \hspace*{\algorithmicindent} Compute probability matrix $Q_+^{(0)} \in \mathbf{(0,1)}^{N^+ \text{x} K}$ with \\ \hspace*{\algorithmicindent} 
 soft K-Means only on positive samples (i.e., $y=+1$)
\State \hspace*{\algorithmicindent} Regularize $Q_+^{(0)}$ with Sinkhorn-Knopp (SK).
\State \hspace*{\algorithmicindent} Apply a Soft$\rightarrow$Hard linear interpolation: \\ \hspace*{\algorithmicindent} \hspace*{\algorithmicindent} \hspace*{\algorithmicindent} $Q_+^{(0)} = \frac{T-t}{T} \text{OneHot}(Q_+^{(0)}) + \frac{t}{T} Q_+^{(0)} $
\State \hspace*{\algorithmicindent} Set $Q_-^{(0)}= \frac{1}{K}$ for background samples ($y=- 1$).
\State \hspace*{\algorithmicindent} Concatenate $Q_-^{(0)}$ and $Q_+^{(0)} $ to get $Q^{(0)} \in \mathbf{(0,1)}^{N \text{x} K}$.
\State \textbf{for} t in T epochs:
   	\State \hspace*{\algorithmicindent} {\bf M step} (supervised step):
       \State \hspace*{\algorithmicindent} Freeze $Q^{(t)}$ 
       \State \hspace*{\algorithmicindent} \textbf{for} expert k in K, estimate:
       \State \hspace*{\algorithmicindent} \hspace*{\algorithmicindent} $p_{\theta, \phi_k}(y_i=+1| x_i, c_i=k)=S(h_{\phi_k}(f_\theta(x_i)))$
       % \State \hspace*{\algorithmicindent} \textbf{end for} 
       \State \hspace*{\algorithmicindent} Estimate $p_{\theta, \psi}(c_i | x_i) = \sigma(h_{\psi}(f_\theta(x_i)))$.
   	\State \hspace*{\algorithmicindent} Compute $L_{\text{experts}}$ and $L_{\text{clustering}}$ (Eq.~\ref{eq:4}).
	\State  \hspace*{\algorithmicindent} {\bf E step} (unsupervised step):
	\State \hspace*{\algorithmicindent} Freeze $\theta^{(t)}$.
   	\State \hspace*{\algorithmicindent} Estimate $Q^{(t+1)}$ as in initialization step.
% \State \textbf{end for}
% \State Return parameters $\theta$, $\phi$ and $\psi$
\end{algorithmic}
\caption{Deep UCSL pseudo-code}
\end{algorithm} 

\newpage

\section{Additional Results}
\subsection{Qualitative Results on Pneumonia and MNIST Experiment}
\begin{figure}[!ht]
    \centering-
    \includegraphics[width=.99\linewidth]{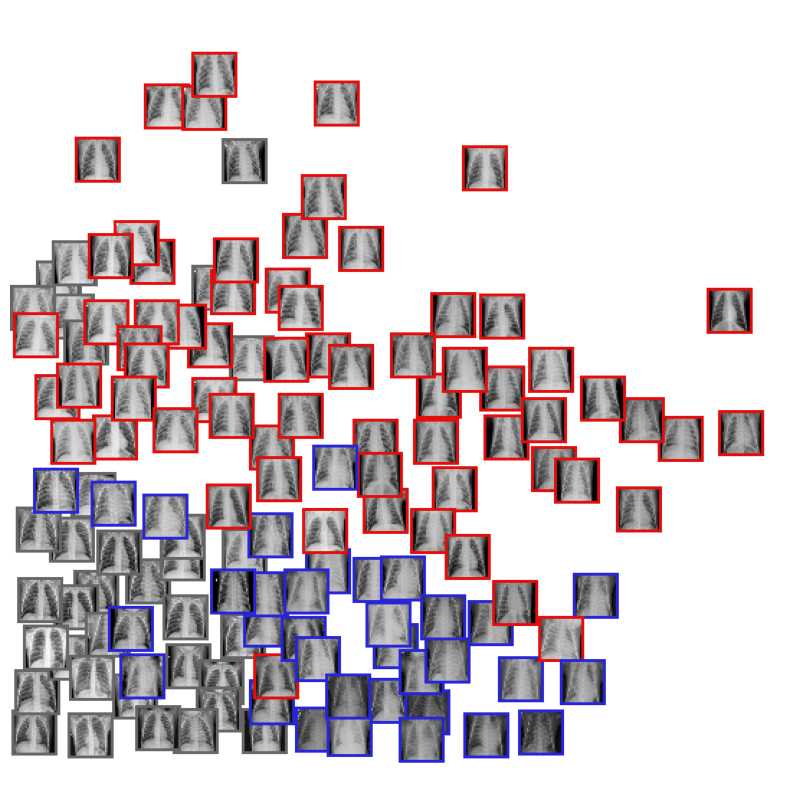}
    \caption{X-Ray pneumonia subtypes. 2D PCA plots of the representation spaces learned by Deep UCSL when seeking two subgroups for X-Ray images of individuals with pneumonia (either bacterial or viral). Red (resp. blue) framed images indicate viral (resp. bacterial) pneumonia. Gray frame images indicate healthy subjects.
    PCA is trained on the TRAIN set and then used on the TEST set for visualization.}
    \label{fig:pneumonia_qualitative_sutyping_examples}
\end{figure}

\begin{figure*}[!tbp]
    \centering
    \includegraphics[width=1\textwidth]%{LaTeX/figures/umap_evolution_across_epochs_neat.png}
    {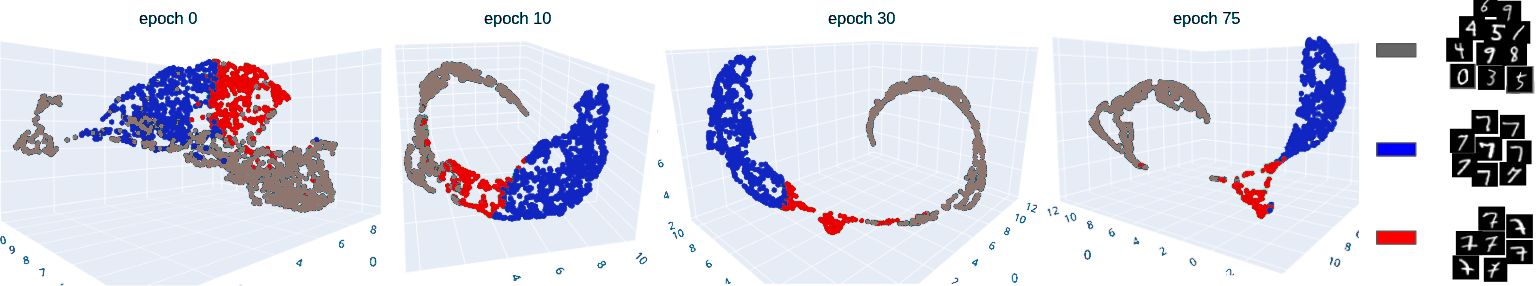}
    \caption{3D UMAP plots of the representation space of Deep UCSL along the epochs on the subgroup identification of the MNIST digit 7. Red and blue points represent the actual (manually labeled) subgroups of the positive class, i.e.: the digit 7 with and without the crossing middle bar. Gray points are samples from the negative class, i.e., other digits than 7. At epoch 1, both classes are fused, there is no clear distinction between subgroups (B-ACC=56\%). During training, classes are progressively separated. At the end of the training (B-ACC=97\%), the two subgroups (red and blue) are well separated and the negative representations (gray) are equidistant from both subgroups. The ratio of distances between healthy and subgroups centroids ($\frac{d(\text{healthy centroid}, \text{subgroup 1 centroid})}{d(\text{healthy centroid}, \text{subgroup 0 centroid})}$) gets close to 1 (equidistant case) during the training (epoch 1: 3.68; epoch 75: 1.35).
    }
    \label{fig:umap_evolution}
\end{figure*}

In Fig.~\ref{fig:pneumonia_qualitative_sutyping_examples}, we provide a qualitative 2D PCA plots of the representations space of Deep UCSL, we observe a clear  separation between the annotated subgroups. In Fig. \ref{fig:umap_evolution}. We observe that both bold and thin digits of the negative class (i.e., all digits but the 7) are encouraged to be equidistant from the subgroups of the digit 7 in the representation space across the epochs.

\subsection{Comparison with weakly supervised deep clustering and foundation models}
In this subsection, we propose to compare Deep UCSL with a naive weakly-supervised deep clustering method and medical imaging foundational models. The naive weakly-supervised deep clustering consists of alternating 1) pseudo-labels estimation (using SK-regularized K-Means), and 2) weights update using a Cross-Entropy loss between clustering prediction and pseudo-labels. The weak supervision comes in during the pseudo-labels estimation, where the healthy/disease labels are concatenated to the subgroup pseudo-labels. Precisely, pathological subgroups are estimated by fitting SK-regularized K-Means on pathological samples only, to produce a pseudo-label matrix of size $N, K$ where $N$ is the number of training samples, and $K$ is the number of pathological subgroups. The diagnosis labels are then concatenated, and the probability of pathological subgroups is set to $0$ for healthy samples. Therefore, the final pseudo-labels matrix has a shape of $N, K$, healthy samples have a pseudo-label one-hot vector equal to $[0, ..., 0, 1]$, and disease samples have a pseudo-label one-hot vector equal to $[Q(c_i=1), ..., Q(c_i=K), 0]$. In our Results Section, we compare the previously described naive method and Deep UCSL and show that Deep UCSL outperforms it on 4 different datasets: MNIST 7 subgroup discovery, viral / bacterial classification of pneumonia, and two eye pathological subgroup discovery: ODIR and RetinalOCT. \\
\noindent Besides, we also compare Deep UCSL with 3 foundation models (FM) trained on large retinal pathology datasets (OCT scans and Retinal Fundus Images), and chest X-Rays datasets in a self-supervised fashion. 
The foundation model we used for Retinal OCTs and fundus images was obtained from \cite{zhou_2023}, the model consisted of a large ViT model \cite{dosovitskiy_2021}, pretrained with Masked Auto Encoder (MAE) technique  on 1.6 million retinal images. The foundational model used for X-Ray images was obtained from \cite{yao_2024}. This model consisted of a Small EVA model \cite{fang_2023}, pretrained with Masked Image Modeling technique  \cite{xiao_2023} encompassing more than 20 distinct human chest health conditions. \RL{Interestingly, despite the "foundamental" objective of EVA-X, we observed poor performances transferability on our experimental setup. We attribute it to the fact that our experimental setup is a pediatric application, which may be under-represented in the EVA-X training set.}
To compare with Deep UCSL, we use the embeddings produced by the foundational models as inputs for K-Means clustering (for Subgroup Discovery) and logistic regression classification (for diagnosis binary classification). Deep UCSL outperforms the Foundational Models (FM) on three medical modalities (X-Ray, Retinal OCT and eye fundus images). We argue that, similarly to contrastive learning techniques such as SimCLR, SCAN, or BYOL, foundation models are trained to encode all latent generative factors in the representation space. As Deep UCSL is trained to ignore healthy factors, while identifying subgroups only based on pathological factors of variability, it outperforms FMs in identifying pathological subgroups. Nevertheless, one could also employ a foundation model as encoder in Deep UCSL, which could be fine-tuned during the optimization process. We leave this exciting perspective for future works.

\section{Implementation Details}
For each dataset, we use different and appropriate architectures, with relative hyper-parameters, like the batch size, that performed well in previous works (more details in each section). The only hyper-parameter proper to Deep UCSL is the Sinkhorn-Knopp strength $\epsilon \in \mathcal{R}^+$. We use the GPU implementation of \cite{caron_2020}, with $\epsilon=0.05$. The hyper-parameter $\epsilon$ controls the strength of the SK regularization. Typically, $\epsilon=0$ means no regularization and $\epsilon\gg0$ brings to subgroups of similar size (i.e.: of a similar number of samples). This is desirable in the medical subgroup discovery context where one often has \textit{a priori} knowledge of the relative size of the subgroups. In practice, we found $\epsilon=0$ to be robust in all our experiments.

\subsection{Neuro-psychiatry application: implementation details.}
\noindent Voxel-based morphometry (VBM) is performed with CAT12 \cite{gaser_2022} to preprocess the images. 
The analysis pipeline includes non-linear spatial registration on a standard template (MNI), Gray Matter (GM), White Matter (WM), and Cerebrospinal Fluid (CSF) tissue segmentation, bias correction and segmentations modulation. 
VBM images are made isotropic with 1.5mm$^3$ spatial resolution, and output dimension is $121 \times 145 \times 121$. 
From there, images are cropped to $121 \times 121 \times 121$ and padded to reach a dimension of $128 \times 128 \times 128$. 
Voxel values are centered on a unit-gaussian distribution per image (\textit{i.e:} mean of voxels of $0$, standard deviation of voxels of $1$). With CAT12, we further computed the Gray Matter (GM) volumes averaged on the Neuromorphometrics atlas that includes 142 brain cortical Regions-of-Interest (ROIs). For Deep Learning methods, we use the pre-processed GM-only images as inputs to a 3D-DenseNet deep encoder, as in \cite{dufumier_2021}. 
For the linear method entitled UCSL, we use GM ROIs features as inputs.

\noindent To compare with an upper bound, we train a Deep Neural Network to classify between SZ and BD in a fully supervised manner with a Binary Cross-Entropy (BCE). For all neuro-psychiatric deep methods, including Deep UCSL, we chose a batch size of $8$, and the data augmentation strategy is similar to \cite{dufumier_2021}: horizontal flip ($p=0.5$); Gaussian blur ($p=0.5$, $\sigma=(0.1, 0.1)$) and noise ($p=0.5$, $\sigma=(0.1, 0.1)$); CutOut ($p=0.5$, patch size $32 \times 32 \times 32$); RandomCrop ($96 \times 96 \times 96$) with $p=0.5$. Deep learning methods were trained with an Adam optimizer, learning rate $10^{-5}$ during $100$ epochs.

\subsection{MNIST experiment: implementation details.}
\noindent For each dataset, we first train on 12,530 MNIST digits (half positive digit (e.g.: 7 or 1) and half the other digits, equivalently distributed, contrarily to \cite{louiset_2021}). In the MNIST experiment, we can notice that boldness is an important general source of variation (see Fig.~\ref{fig:mnist_digits_sutyping_examples}), common among all MNIST digits and thus irrelevant for subgroup discovery. 
In order to encourage boldness invariance, we propose to simulate morphological augmentations (erosion and dilation) during the training of contrastive learning methods (\textit{Morpho} SimCLR and \textit{Morpho} SupCon). This design choice improves the representation quality for subgroup identification, demonstrating that a representation space invariant to general, irrelevant sources of variations provides better features for subgroup discovery. However, this design choice is probably not enough, and we can observe that \textit{Morpho} SimCLR and \textit{Morpho} SupCon performances are inferior to Deep UCSL's. Please note that we also use geometric transformations for every method (Deep UCSL included): 
a RandomRotation with $\pm 25$ degrees, a RandomAffine with translate parameters of $(0.1, 0.1)$, and a shear parameter set to $0.1$.

\noindent Most deep learning methods were trained with an Adam optimizer, a learning rate of $10^{-5}$, trained during $75$ epochs, and a batch size of $256$.  SimCLR and SupCon were trained with Adam, a cosine learning rate schedule, and a batch size of $512$ for $75$ epochs. Deep UCSL relies on a deep convolutional features extractor composed of $4$ convolutional layers with $7 \times 7$ kernels, a padding of $3$, batch normalization layers between each convolution, and numbers of channels equal to $16$, $32$, $64$, and $128$, followed by an average pooling layer producing a representation of size $128$.

\subsection{Pneumonia experiment: implementation details.}
\noindent For the training set, we choose a balanced subset of $1341$ viral samples, $1341$ bacterial samples, and $1341$ healthy samples. For this application, we use the same architecture and image sizes for every method (except for VAE, where we chose an experimental setup similar to \cite{joy_2021}), namely, a ResNet-18 (pre-trained on ImageNet) and $224^2$ pixels images. We trained the methods for $50$ epochs, with a batch size of $256$, and an Adam optimizer with a learning rate $10^{-5}$.

\subsection{Retinal pathology experiments: implementation details}
\noindent For both experiments, we use the same setup for every method (except for VAE, where we chose an experimental setup similar to \cite{joy_2021}), namely, a ResNet-18 (pre-trained on ImageNet) and $224^2$ pixels images. For the Retinal Pathology OCT dataset, the train set comprises 3,000 healthy and 3,000 diseased eye images divided homogeneously into three disease subgroups: Choroidal Neo-Vascularization (CNV), Diabetic Macular Edema (DME), and Drusens in age-related macular degeneration. The ODIR dataset contains 1,890 healthy and 1,391 diseased patients, divided heterogeneously into five disease subgroups. For both experiments, all methods were trained with an Adam optimizer, a learning rate of $10^{-5}$, $50$ epochs, and a batch size of $256$.

\end{appendices}

\end{document}